%% file: main.tex
\documentclass[sigconf]{acmart}
\AtBeginDocument{%
  }

\usepackage{float}
\usepackage{threeparttable}
\usepackage{xspace}
\usepackage{graphicx}
\usepackage{multicol}
\usepackage{multirow}
\usepackage{caption}
\usepackage{subcaption}
\usepackage{comment}
\usepackage{seqsplit}
\usepackage{gnuplottex}
\usepackage{pgfplots}
\usepackage{pgfplotstable}
\pgfplotsset{compat=1.18}

\usepackage{xcolor}
\definecolor{c1}{HTML}{E41A1C}  
\definecolor{c2}{HTML}{377EB8}  
\definecolor{c3}{HTML}{4DAF4A}  
\definecolor{c4}{HTML}{FF7F00}  

\setcopyright{rightsretained}
\copyrightyear{2026}
\acmYear{2026}
\acmConference[CIKM '26]{The 35th ACM International Conference on Information and Knowledge Management}{November 07--11, 2026}{Rome, Italy}

\begin{document}

\input{commands}


\title{GRAG: Generic Response-Augmented Generation Framework for Personalized Conversational Systems}

\author{Junfeng Liu}
\orcid{1234-5678-9012}
\affiliation{%
  \institution{North Carolina State University}
  \city{Raleigh}
  \state{NC}
  \country{USA}
}
\affiliation{%
  \institution{Lirio, Inc.}
  \city{Knoxville}
  \state{TN}
  \country{USA}
}
\email{jliu85@ncsu.edu}

\author{Christopher T. Symons}
\affiliation{%
  \institution{Lirio, Inc.}
  \city{Knoxville}
  \state{TN}
  \country{USA}
}
\email{csymons@lirio.com}

\author{Ranga Raju Vatsavai}
\affiliation{%
  \institution{North Carolina State University}
  \city{Raleigh}
  \state{NC}
  \country{USA}
}
\affiliation{%
  \institution{Lirio, Inc.}
  \city{Knoxville}
  \state{TN}
  \country{USA}
}
\email{rrvatsav@ncsu.edu}


\renewcommand{\shortauthors}{Liu et al.}

\settopmatter{printacmref=false} 

\renewcommand\footnotetextcopyrightpermission[1]{}{} 

\pagestyle{plain}

\input{./sections/abstract}




\maketitle


\input{./sections/intro}
\input{./sections/related_work}

\input{./sections/method}

\input{./sections/experiments}
\input{./sections/results}
\input{./sections/ablation}
\input{./sections/conclusion}
\begin{acks}
The authors express their gratitude to the behavioral scientists for their expert guidance on the design of the human evaluation framework. We thank the internal reviewers for their constructive feedback on early drafts of this paper, and extend our appreciation to the anonymous reviewers for their insightful comments that will help improve the final version of this work. Finally, we thank all the volunteers for their valuable time and dedication during the human evaluation studies.
\end{acks}


\appendix

\input{sections/appendix}





\newpage 

\section*{GenAI Usage Disclosure}

In this work, the authors used generative AI tools as follows:
\begin{itemize}
    \item LLMs were used to generate generic responses (Section~\ref{sec:method:generic}). The generic responses are used as auxiliary data to train the models. This is a part of the core contributions of this work. 
    \item LLMs were used in the LLM-as-a-judge evaluation (Section~\ref{sec:results:human-llm-eval}) to provide additional evidence and support of performance of the proposed method. 
    \item AI-based coding agent was used to implement the human evaluation web application (Section~\ref{sec:results:human-llm-eval}). This facilitates the process to collect human annotations, and is not a part of the core contributions of this work. 
    \item Generative AI agent was used for editing, language refinement, and formatting suggestions during the writing of this work.
\end{itemize}

After using the above tools, the authors have reviewed and edited the generated content/code as needed and will take full responsibility for the content of the publication.

\bibliographystyle{ACM-Reference-Format}
\bibliography{main}

\end{document}

%% file: commands.tex
\newcommand{\etal}{\textit{et al.}\xspace}

\newcommand{\preicl}[1][]{\texttt{ICL\if\relax\detokenize{#1}\relax\else(#1)\fi}\xspace}
\newcommand{\presft}[1][]{\texttt{PreSFT\if\relax\detokenize{#1}\relax\else(#1)\fi}\xspace}
\newcommand{\ourmethod}{\texttt{GRAG}\xspace}
\newcommand{\ourmethodpre}{\texttt{PreGRAG}\xspace}
\newcommand{\ourmethodpost}{\texttt{PostGRAG}\xspace}
\newcommand{\real}{\mathbb{R}\xspace}
\newcommand{\aggr}{\texttt{aggr}\xspace}
\newcommand{\context}{\texttt{CTXT}\xspace}
\newcommand{\candidate}{\texttt{CAND}\xspace}
\newcommand{\softmax}{\texttt{softmax}\xspace}
\newcommand{\simi}{\texttt{sim}\xspace}
\newcommand{\identity}[1]{\mathbb{I}({#1})\xspace}

\newcommand{\stylevector}{\mathcal{S}}
\newcommand{\styleembedder}{\text{SE}}

%% file: sections/abstract.tex
\begin{abstract}

Deploying highly capable personalized conversational agents in resource-constrained or privacy-sensitive environments remains a significant challenge. We identify a fundamental bottleneck in the existing approaches: current training paradigms treat personalization and grounding as a single monolithic learning problem. 
Under these paradigms, language models are forced to simultaneously address \textit{what} to say (content grounding) and \textit{how} to say it in a user-specific way (personalization), which introduces significant computational and optimization challenges. Consequently, contextual grounding is often sacrificed for persona adherence, or vice versa, resulting in responses that are either weakly grounded in the conversational history or insufficiently personalized.
%
%
In this work, we propose the Generic Response-Augmented Generation (\ourmethod) framework that decouples these competing objectives by leveraging offline, generic responses from high-capacity, general-purpose LLMs as a semantic and structural scaffold to guide the fine-tuning of smaller, task-specialized models seamlessly in resource-limited environments. 
By decoupling the content grounding from personalization, \ourmethod allows the model to focus exclusively on persona injection while remaining firmly anchored to the conversational context.
We instantiate the \ourmethod in two post- and pre-fusion-based architectural variants and evaluate them on multiple benchmark conversational datasets that cover diverse personalization structures. 
Our results demonstrate that \ourmethod significantly outperforms state-of-the-art methods that do not use auxiliary scaffolding, yielding up to 47\% improvements in ROUGE-2 and 36\% in BLEU scores. 
Ultimately, \ourmethod offers a generalizable blueprint for building grounding-aware personalized conversational systems in resource-limited environments.

\end{abstract}

%% file: sections/intro.tex
\section{Introduction}
\label{sec:intro}

Personalized conversational systems have become increasingly important for applications where personalized, persuasive, or empathetic responses are required, such as virtual assistants, AI companions, and healthcare support. A key challenge in this setting is for the conversational systems to generate responses that are not only consistent with and tailored to the user-specific profiles or attributes (personas), but also well-grounded in the conversational context. In practice, these two requirements are tightly coupled: failures in grounding often lead to irrelevant responses, while lack of personalization results in generic and bland responses.

Recent progress in massive large language models (LLMs) has enabled powerful conversational agents such as ChatGPT~\cite{achiam2023gpt} and Claude~\cite{anthropic2025claude4}. However, these agents face several practical challenges in resource-limited, real-world applications.
First, they are optimized for broad task coverage rather than specialized personalization outcomes.
Second, they introduce third-party privacy and security concerns in sensitive domains like healthcare.
Third, their massive scale (often hundreds of billions of parameters) makes task-specific fine-tuning and deployment prohibitively expensive.
%
Smaller LLMs are therefore more practical for fine-tuning and real-world deployment~\cite{hu2022lora, touvron2023llama}. Yet, effectively integrating conversation history and persona information into these models remains challenging. State-of-the-art (SOTA) approaches~\cite{zhang2018personalizing} concatenate these inputs directly into the prompt. This monolithic placement forces the model to simultaneously resolve \textit{what} to say (content grounding) and \textit{how} to say it (personalization), often leading to incoherent or insufficiently tailored responses.

In this work, we depart from the traditional monolithic paradigm and introduce an intermediate semantic scaffold derived from the non-personalized output (referred to  as the ``generic response'') of a high-capacity, general-purpose LLM (referred to  as the ``generic LLM''). Intuitively, while these generic responses may not reflect user-specific attributes or task-oriented tones, the advanced reasoning capabilities of the generic LLM allow the model to accurately capture core conversational intents and discourse structures. Consequently, these outputs provides a robust grounding signal to effectively guide the generation of the final personalized responses.

Based on this intuition, we present the Generic Response-Augmented Generation (\ourmethod) framework. By utilizing offline generic responses as structural scaffolds, \ourmethod offloads the heavy learning burden of contextual alignment from the smaller, task-specialized model (referred to as the ``task LLM''), allowing it to focus exclusively on persona injection. This separation enables local deployment on commodity edge hardware (e.g., consumer-grade GPUs), offering a cost-effective and privacy-preserving alternative to cloud-based APIs. The main contributions of this work are as follows:
\begin{itemize}
    \item We propose a novel \ourmethod framework that leverages auxiliary generic responses as scaffolding to decouple conversational discourse structure from persona injection.
    \item We design and compare pre-fusion and post-fusion architectures to effectively integrate generic response scaffolds with conversational contexts.
    \item We demonstrate consistent and significant improvements over SOTA monolithic baselines across multiple benchmark datasets through comprehensive evaluations and analyses.
    \item We offer new empirical insights revealing that contextual grounding, rather than personalization, is the primary bottleneck for small task LLMs in conversational systems.
\end{itemize}

The rest of this work is organized as follows.
Section~\ref{sec:related} reviews related work.
Section~\ref{sec:method} introduces the pre- and post-fusion-based \ourmethod methods.
Sections~\ref{sec:experiments} and~\ref{sec:results} present the experimental setup and results.
Sections~\ref{sec:ablation} and~\ref{sec:limitations} discuss the ablation studies and the limitations of the \ourmethod method,
followed by a conclusion in Section~\ref{sec:conclusions}.

%% file: sections/related_work.tex
\section{Related Work}
\label{sec:related}

Prior work on personalized conversational systems has largely treated grounding and personalization as a monolithic learning problem. We present our literature review around four different areas to reveal both the limitations of existing approaches and the opportunities that motivate the design of \ourmethod. 

\subsection{Personalized Conversational Systems}
\label{sec:related:personalized}

Personalization has become critical in many real-world conversational systems. For example, in precision nudging~\cite{bucher2023patient}, the agents are expected to leverage the patient personal health profiles and deliver effective and personalized communications across diverse populations. 
Many research efforts have been focused on how to effectively leverage persona data in response generation~\cite{liu2023pcpe, roller2021recipes, humeau2019poly, zhang2018personalizing}. 
Zhang~\etal~\cite{zhang2018personalizing} created the Persona-Chat dataset that incorporates user personal profile information in social conversations, which was later expanded to the benchmark ConvAI2 dataset~\cite{dinan2020second}.
Humeau~\etal~\cite{humeau2019poly} developed a sentence scoring model, Poly-Encoder, that learns global level self-attention features between the conversational context and the candidate responses. Liu~\etal~\cite{liu2023pcpe} proposed Persona-Coded Poly-Encoder that improves the sentence scoring of Poly-Encoder by leveraging two specialized streams for persona and conversational context. These methods consider the conversational problem as response selection from a set of pre-defined candidates. These selection-based approaches typically ensure good quality of the responses, while they also suffer from response novelty issues, which makes generation-based approaches more popular in various applications~\cite{roller2021recipes, liu2020you, zhang2020dialogpt}. Liu~\etal~\cite{liu2020you} created a mutual persona perception mechanism to model how each speaker perceives the other. 
%
These existing methods fail to leverage valuable auxiliary resources and rely solely on signals from the given dataset, making the models sensitive to even minor quality issues in the data.

\subsection{Auxiliary Learning Signals for Natural Language Generation}
\label{sec:related:auxiliary}

A broader line of research leverages auxiliary learning signals to guide the NLG process.
Several efforts incorporate auxiliary signals through multi-task learning.
For example, Rothe~\etal~\cite{rothe2020leveraging} leveraged publicly available model checkpoints as auxiliary signal to warm-start model training.
A particular stream of research is interested in leveraging external knowledge and grounding signals that help align the responses with knowledge and facts~\cite{liu2025comac, jang2022call, dinan2018wizard}, visual environment~\cite{mostafazadeh2017image}, or other constraints. For example, 
Jang~\etal~\cite{jang2022call} developed a conversational dataset with both persona and knowledge for personalized conversations that adheres to facts and knowledge, and Liu~\etal~\cite{liu2025comac} proposed post-fusion-based grounding networks with sparse token similarity to improve the persona and knowledge grounding quality. 
However, these methods typically require carefully curated auxiliary sources (e.g., well-constructed knowledge base) that are relevant to the main task (e.g., knowledge-adherence conversation), which often requires domain expertise for collection and processing before the model could learn. 
RAG systems~\cite{lewis2020rag} are also widely adopted in both academic and industrial settings by retrieving a subset of relevant documents from a vector database based on document and query similarities and use those retrieved documents as auxiliary signal for LLMs. Yet, RAG systems also suffer from several limitations such as dependencies on the qualities of the selected document chunking and encoding strategies, as well as scalability issues with large scale documents.

\subsection{Knowledge from Large Language Models}
\label{sec:related:llm_knowledge}
Compared to the human curated auxiliary sources, strong pre-trained LLMs offer economical alternatives that create useful learning signals at scale. A typical application is via in-context learning (ICL)~\cite{dong2024survey, wang2023large, reif2022recipe}, which 
involves prompting a pre-trained LLM at inference time with few-shot examples and a task instruction.
For example, Wang~\etal~\cite{wang2023large} leveraged LLMs to detect implicit topics to select demonstrations that share similar topics with the target input under ICL framework. 
Another popular way to leverage knowledge from LLMs is distillation~\cite{hinton2015distilling}, which fine-tunes a smaller LLM with the examples generated by a larger pre-trained LLM. 
However, the generation outputs are entirely dependent on the quality of the pre-trained LLMs, and errors or hallucinations may be inherited when the model's knowledge is flawed.

\subsection{Text Style Transfer}
\label{sec:related:tst}

Text style transfer (TST) approaches~\cite{hu2022text} involve converting a text from one style into another, and are widely used in style recreation~\cite{pan2024unsupervised}, paraphrase~\cite{yu2023language, jia2025syntax}, etc.
Traditional TST approaches~\cite{lample2019multiple, hu2017toward} typically disentangle the style and semantics of a source text, then convert it to a target text in a target style while maintaining the semantics. Further explorations have been made towards TST problems without explicit disentanglement~\cite{dai2019style}. Recent methods attempt to leverage pre-trained LLMs using ICL~\cite{reif2022recipe} or distillation~\cite{zhang2024distilling} training paradigms. 
%
%
However, pure TST approaches often fail in conversational settings because they solely rely on the parallel source and target text pairs, ignoring valuable context information such as conversation histories and discourse.  
There have been research attempts on adapting TST approaches to conversational problems via RNN~\cite{zhang2019neural} and variational auto-encoders~\cite{saha2022stylistic}. However, these methods often require multiple parallel responses with the same semantics but in different styles. In addition, they rely heavily on human crafted style labels. These challenges make it difficult to adjust TST approaches to real applications with various, dynamic persona data and non-parallel responses.

%% file: sections/method.tex
\input{figures/model_overview}

\section{Method}
\label{sec:method}

In this work, we train a conversational model that is able to generate personalized responses to a user question by leveraging  generic responses of a generic LLM. 
The generic response generation process can be viewed as a highly lossy, unrecoverable compression of the conversational history. Therefore, instead of relying on a traditional TST approach, we design \ourmethod as a multi-source input framework that processes both the conversational history and persona while leveraging generic responses as auxiliary guiding signals.
Given a user's text-based persona profile $P$, conversation history $H$ that ends with a user utterance, and an augmented generic response $R_G$ generated by a generic LLM, the goal is to train a model to generate a personalized response $R_P$ for the conversation. 
%

The main intuition is that $R_G$'s carry basic structure and semantics of the language, while they may lack necessary personalization elements to make the responses tailored to specific users (referred to as ``style'' in this work). Therefore, the problem can be framed as two questions:
1) how to learn the personalization style for a specific user,
and 2) how to leverage this personalization style to enhance the underlying semantics.
%
Fusing these multi-source inputs ($P$, $H$, and $R_G$) spans an architectural design spectrum from raw input concatenation (pre-fusion)~\cite{zhang2018personalizing} to intermediate layer routing (mid-fusion)~\cite{humeau2019poly} and isolated modular representation (post-fusion)~\cite{liu2022persona}. To model the style and personalization, we instantiate \ourmethod using the two boundary extremes of this spectrum: a pre-fusion architecture that learns via implicit self-attention, and a post-fusion architecture that learns via explicit decoupling and dedicated style embeddings.
Figure~\ref{fig:our-approach} shows the pre-fusion and post-fusion variations of \ourmethod.
%
Specific choices of LM backbone will be discussed in Section~\ref{sec:experiments:setup:language_model}.

\subsection{Generic Response Augmentation}
\label{sec:method:generic}
Given a base conversational dataset containing conversational  histories $H$ and the original agent responses $R_P$, which are typically personalized, we first augment the dataset by generating generic responses $R_G$ using a generic LLM configured with the  prompt templates detailed in Appendix~\ref{appendix:prompts}. In the system prompts, we provide a description of the task and the nature of the conversation, along with rules that instruct the LLM to generate responses with a neutral style and emotional tone. Responses are generated for each non-leading \texttt{<agent>} turn in a conversation. In the user prompts, we provide the conversational  history $H$ with the last utterance of the  \texttt{<user>} role isolated. During the generation process, the persona $P$ is withheld from the LLM (unless explicitly discussed in $H$). The resulting outputs  are then utilized as $R_G$.

\subsection{Pre-Fusion-Based Method}
\label{sec:method:pre-fusion}

We denote the pre-fusion-based architecture as \ourmethodpre. A typical pre-fusion approach is to concatenate raw inputs before modeling. Here  $P$, $H$, and $R_G$ are concatenated as a single long input, $H' = [P; H; R_G]$, for a transformer-based task LLM.
For encoder-decoder models, $H'$ is encoded as $E_{H'} = Encoder(H')$ and $E_{H'}$ is the encoder hidden state for the decoder. For decoder-only models, $E_{H'} = Decoder(H')$. 
The decoder output represents the logits for the personalized response $R_P$.

\subsection{Post-Fusion-Based Method}
\label{sec:method:post-fusion}

Many existing studies have demonstrated the better performance of post-fusion-based methods over pre-fusion-based methods, especially for heterogeneous inputs~\cite{liu2023pcpe}, as the models are able to learn dedicated signals from each source. Here we consider a post-fusion architecture, denoted as \ourmethodpost, that encodes multiple input sources of the conversational context separately and enhances the response generation through an explicit style embedding. 

In \ourmethodpost, the persona $P$, conversation history $H$ and the generic response $R_G$ are independently encoded by a transformer-based encoder as $E_{P} = Encode(P), E_{H} = Encode(H), E_{R_G} = Encode(R_G)$. 

\subsubsection{Personalization Style Extractor}
\label{sec:method:style-encoder}
A personalization style embedding $\stylevector_P$ is computed by an attention-based extraction layer $\styleembedder_P$, which derives a contextual representation of the persona conditioned on the conversational context. 
Specifically, $\styleembedder_P$ takes the persona $E_{P}$ as the query and the concatenation of $[E_{H}; E_{R_G}]$ as the key and value for the attention layer.
A masked mean pooling then reduces the sequence length dimension to produce the final style embedding $\stylevector_P$. 
Intuitively, $\stylevector_P$ represents the information discrepancy between the persona and the other inputs in an abstract embedding space (i.e., what information is missing in the history and the generic response to produce a personalized response). Such discrepancy is treated as the essential personalization style signal for the subsequent personalized response generation. 

\subsubsection{Style Enhanced Decoder}
\label{sec:method:style-decoder}
To decode the conversational context into a personalized response $R_P$, we adopt a transformer-based Style Enhanced Decoder. Specifically, the history embedding $E_{H}$ is used as the encoder hidden state, and the decoder output represents the logits for $R_P$.
Within the decoder, the personalized style embedding $\stylevector_P$ supplements the hidden states at each decoding layer, i.e., $h_i \leftarrow \stylevector_P + h_i$, where $h_i$ is the hidden state of the $i$-th decoder layer.

\subsubsection{Style Removal Regularization}
\label{sec:method:style-regularization}
We observe that while style enhancement could largely improve the personalization quality for most of the responses, it could also introduce unnecessary noise into certain types of responses where personalization is not critical (e.g., greetings or objective questions). 
To mitigate this issue introduced by unnecessary style enhancement, we consider removing stylistic information from the response. This style removal regularization ensures the model can disentangle the semantics and personalization so that the fundamental semantics (represented by the generic response) can be reconstructed. 

In this step, \ourmethodpost learns a null style vector, $\stylevector_G$, that represents a hypothetical ``generic style'' without any personalization, and then back-translate the personalized responses $R_P$ to the generic response $R_G$. 
We employ a separately trained generic style extractor, $\styleembedder_G$, that shares a similar architecture to $\styleembedder_P$ to learn the null style vectors $\stylevector_G$. $\styleembedder_G$ is an attention learning layer that takes $E_{P}$ as the query and the concatenation of $[E_{H}; E_{R_P}]$ as the key and value. 
Intuitively, by using the $P$ as a query to identify user-specific elements within the $H$ and $R_P$, the model learns to isolate and factor out these traits, leaving behind a residual representation of the generic style.
In the back-translation stage, $R_P$ is used as input to the style enhanced decoder, with the null style vector $\stylevector_G$ supplementing at each decoder layer. The output of the decoder represents the logits of reconstructing $R_G$.

\subsection{Contrastive JEPA Learning}
\label{sec:method:jepa}

With the goal of conversation personalization, we further adopt a contrastive JEPA (C-JEPA) learning objective based on the Joint-Embedding Predictive Architecture (JEPA)~\cite{assran2023self}. 
While standard language modeling losses (e.g., next word prediction) treat all plausible responses equally, they do not explicitly encode the distinctions between $R_P$ and $R_G$. The proposed C-JEPA naturally bridges this gap by treating  $R_G$ as a negative target in the joint embedding space, allowing the model to explicitly learn what makes a response personalized rather than merely  fluent.
%
%
Specifically, $R_P$ is encoded as $E_{R_P} = Encode(R_P)$. We use the last non-padding token embedding of $E_{R_P}$ and $E_{R_G}$ as the representation of personalized and generic responses, denoted as $\mathcal{E}_{P}$ and $\mathcal{E}_{G}$, respectively. For the context representation $\mathcal{E}_{C}$, we use the last non-padding token of $E_{H'}$ and  $E_{H}$ for \ourmethodpre and \ourmethodpost, respectively. We use cosine distance as the metric when comparing embeddings, following existing work that leverages JEPA training~\cite{huang2025llmjepa, chen2025vl-jepa}. The distances between $\mathcal{E}_{C}$ and $\mathcal{E}_{P}$ and between $\mathcal{E}_{C}$ and $\mathcal{E}_{G}$ are computed as 
$\mathcal{D}_{CP} = 1 - cosine(\mathcal{E}_{C}, \mathcal{E}_{P})$ and $\mathcal{D}_{CG} = 1 - cosine(\mathcal{E}_{C}, \mathcal{E}_{G})$,
and the C-JEPA loss is 
\begin{equation}
\label{eqn:loss_jepa}
    L_{C-JEPA} = \mathcal{D}_{CG} - \mathcal{D}_{CP} = cosine(\mathcal{E}_{C}, \mathcal{E}_{P}) - cosine(\mathcal{E}_{C}, \mathcal{E}_{G}).
\end{equation}
Notably, when $R_P$ and $R_G$ are semantically similar (e.g., in the case of objective or factual responses), $\mathcal{E}_{P}$ and $\mathcal{E}_{G}$ will naturally be similar, causing $L_{C-JEPA}$ to approach zero. This ensures that the C-JEPA loss does not impose a spurious penalty on the model.

\subsection{Training and Inference}
\label{sec:method:training}

For training, we use aggregated token-level cross-entropy loss based on decoder logits as the base language modeling objective for next-word prediction (NWP), combined with other applicable loss terms.

\textbf{For the pre-fusion \ourmethod method}, the model is optimized using the following loss function:
\begin{equation}
\label{eqn:loss_pre}
L_{pre} = L_{P} + \beta L_{C-JEPA}.
\end{equation}

\textbf{For the post-fusion \ourmethod method}, the model is optimized by:
\begin{equation}
\label{eqn:loss_post}
L_{post} = L_{P} + \alpha L_{G} + \beta L_{C-JEPA}.
\end{equation}

In Eq.~\ref{eqn:loss_pre} and~\ref{eqn:loss_post}, 
$L_{P}$ is the NWP loss of personalized response,
$L_{G}$ is the NWP loss of the style removal regularization,
$L_{C-JEPA}$ is the C-JEPA regularization loss from Eq.~\ref{eqn:loss_jepa},
and $\alpha$/$\beta$ are two weighting factors for the applicable loss terms. 
Note that $L_{G}$ is not applied to \ourmethodpre, since handling style removal without a dedicated component in pre-fusion architecture imposes extra learning complexity to the model, leading to suboptimal performance.

For inference, a sequence of tokens is sampled as the response based on the NWP logits in an auto-regressive framework~\cite{hoogeboom2021autoregressive}. Note that both the style removal regularization and contrastive JEPA regularization are applied only during training, when the personalized response $R_P$ is available.

%% file: figures/model_overview.tex
\begin{figure*}[ht]
    \centering
    
    \begin{subfigure}[ht]{0.43\textwidth}
        \centering
        \includegraphics[height=3.8cm]{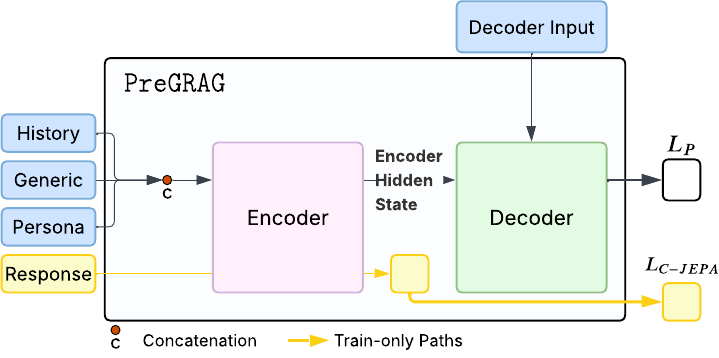}
        \caption{Pre-Fusion-Based \ourmethodpre. The inputs are concatenated as a long input before modeling.}
    \end{subfigure}
    \hfill
    \begin{subfigure}[ht]{0.55\textwidth}
        \centering
        \includegraphics[height=3.8cm]{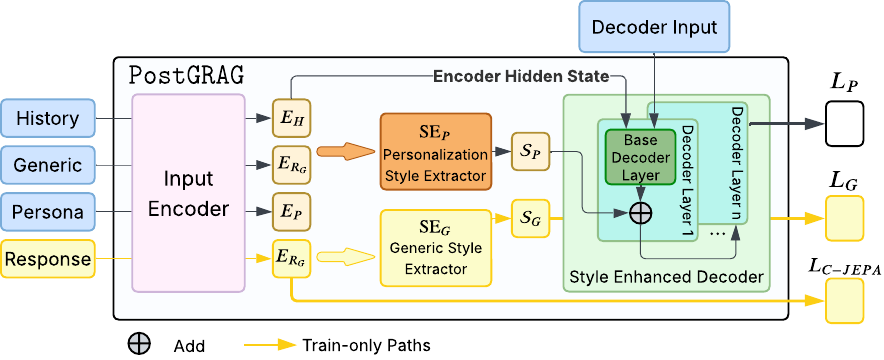}
        \caption{Post-Fusion-Based \ourmethodpost. The inputs are encoded separately, and an explicit style embedding $\stylevector_P$ is learned and used in the style enhanced decoder.}
    \end{subfigure}
    \caption{Overview of \ourmethod Methods}
    \label{fig:our-approach}
\end{figure*}

%% file: sections/experiments.tex
\section{Experiments}
\label{sec:experiments}

\subsection{Training Tasks and Base Datasets}
\label{sec:experiments:datasets}

To evaluate our method, we consider three benchmark conversational datasets that cover various types of personalization traits. In this work, we may use ``tasks'' and ``datasets'' interchangeably. 


%
\textbf{Synthetic Persona Chat (SPC)}~\cite{jandaghi2024faithful} is a synthetic social chit-chat dataset that addresses the faithfulness and toxicity issues present in the original ConvAI2 dataset~\cite{dinan2020second}. In SPC, an LLM generates text-based persona profiles free of inherent inconsistencies, assigns unique  persona profiles to two speakers, and  simulates a conversation between them to get to know each other.

\textbf{EmpatheticDialogues (ED)}~\cite{rashkin2019towards} is a conversation dataset that incorporates empathetic feelings into the conversation partner's responses. Each conversation has an empathetic label that indicates the specific empathetic trait of the conversation. The empathetic labels can be treated as simple persona attributes that reflect the speaking style of the conversation partners. 

\textbf{Persuasion-for-Good (PFG)}~\cite{wang2019persuasion} is a persuasive conversation dataset where agents (persuaders) try to persuade users (persuadees) to make donations to a charitable organization. The agents have access to the user's persona profile, including nine basic demographic attributes and twenty-three psychological assessment scores from user surveys. The conversations reflect personalized persuasion strategies based on the user profiles. 

\input{tables/dataset_statistics}

For datasets without separate validation and test sets, we randomly split 90\% / 10\% of the conversations for training / validation.
Table~\ref{table:dataset} shows the statistics of the three task datasets. 
Table~\ref{table:generic_example} shows examples of different types of generic responses generated for the PFG and SPC datasets, demonstrating what are the added values and what are missing.
In Ex. 1, $R_G$ captures the core semantics around ``reuniting families'', but it uses a flat, unempathetic tone and fails to leverage the ``caring female'' traits from persona.
In Ex. 2, 
$R_G$ anchors the main discourse structure ``I'm a \texttt{[position]}'', but it is unable to provide the correct position information since it does not have access to persona data. 
In Ex. 3, $R_G$ provides external knowledge of the topic on a mystery book writer and her books, which is absent from the conversational context and might be helpful for better knowledge-grounded responses.

\input{tables/generic_response_example}


\subsection{Baseline Methods}
\label{sec:experiments:baseline}

In this work, we consider four pre-fusion-based methods across two popular learning schemes as the baselines for comparison.

\subsubsection{Few-shot In-Context Learning (\preicl)}
The \preicl scheme generates responses with a strong pre-trained LLM by learning from examples in the prompt without fine-tuning.
%
We consider two prompting paradigms under the \preicl scheme. 
The first paradigm, denoted as \textbf{\preicl[Hist]}, generates responses in a traditional conversation setting where the inputs consist of persona and conversation history. 
The second paradigm, denoted as \textbf{\preicl[TST]}, generates responses using a TST approach based on persona and generic response, and the model is instructed to rewrite the generic response into a personalized version. 
Appendix~\ref{appendix:prompts} shows the ICL prompts. 

\subsubsection{Pre-Fusion-Based Supervised Fine-tuning (\presft)}
The \presft methods fine-tune smaller task LLMs using pre-fusion architectures to align the model output with $R_P$. We consider the same two paradigms: traditional conversation generation using persona and conversation history, and TST using persona and generic response, denoted as \textbf{\presft[Hist]} and \textbf{\presft[TST]}, respectively.

    
    

\subsection{Experimental Setup}
\label{sec:experiments:setup}

\subsubsection{Implementation\protect\footnotemark}
\footnotetext[1]{The source code and augmented datasets will be made available on GitHub soon.}

All the model initializations are loaded from Hugging Face~\footnote{\url{https://huggingface.co}}.
%
%
For generic response augmentation, we use 4-bit quantized Llama-3.1-8B-Instruct as the generic LLM with tradeoff between performance and efficiency. 
For \preicl baselines, we use 4-bit quantized Qwen2.5-32B-Instruct model~\cite{yang2025qwen3} with the number of shots $n=0\sim5$. When $n>0$, experiments are repeated five times with different random seeds and the reported results are averaged over the five repeats to reduce random bias of training shot sampling. 
For \presft and \ourmethod models, we use BART-base~\cite{lewis2019bart} initialization and fine-tune with AdamW optimizer~\cite{loshchilov2019decoupled} over the training partition of each task for 20 epochs on NVIDIA RTX4060-Ti GPUs with 16GB memory.
For \ourmethod models, we perform a hyper-parameter search over $\alpha \in \{0, 0.1, 0.2, \cdots, 1.0\}$,  $\beta \in \{0, 10^{-3}, 5\times10^{-3}, \cdots, 10^{1}\}$ spanning four orders of magnitude, and initial learning rates $\lambda_0 \in \{10^{-3}, 10^{-4}, 10^{-5}\}$ decaying linearly to 50\% of $\lambda_0$ by the final optimization step.

\subsubsection{Language Model Choices}
\label{sec:experiments:setup:language_model}

We evaluate a range of pre-trained models as architectures and initializations for both \presft and \ourmethod, including encoder-decoder models (BART-base~\cite{lewis2019bart} and T5-base~\cite{raffel2020t5}) and decoder-only models (GPT2-medium~\cite{radford2019gpt2}, Llama3.2-1B-Instruct~\cite{touvron2023llama}, and Qwen3-0.6B~\cite{yang2025qwen3}) with parameter scales suitable  for commodity edge devices. For decoder-only \ourmethod models, the decoder is repurposed to execute the encoding steps.
%
For T5, Llama3.2 and Qwen3, we adopt QLoRA~\cite{dettmers2023qlora} for memory efficiency.

Among all candidates, BART achieves the best overall performance for both \presft and \ourmethod, which is attributable to three factors.
First, encoder-decoder models have been shown to be better suited for tasks requiring summarization and reasoning over multiple heterogeneous inputs~\cite{liu2025comac}.
Second, BART is pre-trained to reconstruct documents via diverse denoising strategies, which confers a natural advantage for generation and summarization tasks. 
Third, its relatively compact size ($\sim$140M parameters) reduces the risk of overfitting, particularly when fine-tuning on small task-specific datasets.
Accordingly, all subsequent experiments and analyses adopt BART as the base architecture and initialization.


\subsubsection{Evaluation Metrics}
We evaluate \ourmethod based on several automated metrics against the predicted and reference responses.
\textbf{Char-F1} measures the character-level overlap between the predicted and reference responses.
\textbf{ROUGE} and \textbf{BLEU} measure discourse structure alignment based on n-gram similarity of the predicted response to the reference. 
Bits-per-Character (\textbf{BPC}) is a rough approximation of the model's entropy over the reference response and measures the model's information compression efficiency. BPC serves as an alternative to Perplexity (PPL), another commonly used metric. Unlike PPL, BPC scores are directly comparable across models with different tokenizers and vocabulary spaces.
BERT precision / recall / F1 scores (\textbf{BERT-P} / \textbf{BERT-R} / \textbf{BERT-F1})~\cite{zhang2020bertscore} measure semantic overlap with the reference. Since they do not explicitly capture persona adherence, we treat them merely as soft indicators of semantic alignment and rely on human and LLM evaluations (Section~\ref{sec:results:human-llm-eval}) for rigorous personalization assessment.

Higher F1 / ROUGE / BLEU / BERT scores and lower BPC (denoted by BPC$^{\downarrow}$) values indicate better performance of a model. 




%% file: tables/dataset_statistics.tex
\begin{table}
\centering
\caption{Dataset Statistics}
\label{table:dataset}


\small
\begin{threeparttable}
\bgroup
\def\arraystretch{1}%
\begin{tabular}{
  @{\hspace{5pt}}l@{\hspace{5pt}}
  @{\hspace{10pt}}r@{\hspace{5pt}}
  @{\hspace{5pt}}r@{\hspace{5pt}}
  @{\hspace{10pt}}r@{\hspace{5pt}}
  @{\hspace{5pt}}r@{\hspace{5pt}}
  @{\hspace{10pt}}r@{\hspace{5pt}}
  @{\hspace{5pt}}r@{\hspace{5pt}}
}
\toprule

     & \multicolumn{2}{c}{Train} & \multicolumn{2}{c}{Validation} & \multicolumn{2}{c}{Test} \\
     \cmidrule(r){2-3} \cmidrule(r){4-5} \cmidrule(r){6-7}
Task & \# Conv.  & \# A.T. & \# Conv.  & \# A.T. & \# Conv.  & \# A.T. \\

\midrule 

SPC      & 8,369  & 112,264 & 941   & 12,894 & 901   & 12,167 \\
ED       & 19,533 & 40,254  & 2,770 & 5,738  & 2,547 & 5,259 \\
PFG\tnote{$\ast$}     & 888    & 8,386   & 98    & 920    & 98    & 920\\

\bottomrule
\end{tabular}

\begin{tablenotes}[flushleft]
    \setlength\labelsep{0pt}
    \footnotesize
    \item ``\# Conv.'' and ``\# A.T.'' represent the number of conversations and the number of agent turns (or data points). The PFG\tnote{$\ast$} dataset shares the same validation and test partitions.
    
\end{tablenotes}
\egroup 
\end{threeparttable}


\end{table}

%% file: tables/generic_response_example.tex
\begin{table}[t]
\centering

\caption{Examples of Augmented Generic Responses}
\label{table:generic_example}

\small
\begin{threeparttable}
\bgroup
\def\arraystretch{1}%
\begin{tabular}{
    @{\hspace{0pt}}p{1.1cm}@{\hspace{5pt}}|
    @{\hspace{5pt}}l@{\hspace{5pt}}
}
\multicolumn{2}{@{\hspace{0pt}}p{\dimexpr 1.1cm + 10pt + 0.8\columnwidth\relax}@{}}{%
    \textbf{Ex. 1. (PFG)}: $R_G$ captures the semantics in a flat tone without persona.
} \\

\toprule
Persona\tnote{$\dagger$} $P$ &  
\begin{minipage}[t]{0.8\columnwidth}%
<user\_persona> age: 29.0, sex: Female, care: 5.6, ...\\
<agent\_persona> age: 37.0, sex: Female, care: 5.0, ...
\end{minipage}
\\
\midrule

History $H$ & 
\begin{minipage}[t]{0.8\columnwidth}%
... <user> that's really good to hear! I can't imagine not being able to see my son!
\end{minipage}
\\
\midrule

Reference Response $R_P$ & 
\begin{minipage}[t]{0.8\columnwidth}%
<agent> Children shouldn't be separated from their families. It is very heartbreaking to hear, and I'm glad Save the Children stepped in.
\end{minipage}
\\
\midrule

Generic Response $R_G$  & 
\begin{minipage}[t]{0.8\columnwidth}%
<agent> That's understandable. Save the Children works towards reuniting families and providing essential services to vulnerable children worldwide, including ...
\end{minipage}
\\

\bottomrule

\end{tabular}
\vspace{5pt}
\begin{tabular}{
    @{\hspace{0pt}}p{1.1cm}@{\hspace{5pt}}|
    @{\hspace{5pt}}l@{\hspace{5pt}}
}

\multicolumn{2}{@{\hspace{0pt}}p{\dimexpr 1.1cm + 10pt + 0.8\columnwidth\relax}@{}}{%
    \textbf{Ex. 2. (SPC)}: $R_G$ anchors the discourse structure without persona data.
} \\

\toprule
$P$ &  
\begin{minipage}[t]{0.8\columnwidth}%
<user\_persona> I am a professional skater...\\
<agent\_persona> I am a stock broker...
\end{minipage}
\\
\midrule

$H$ & 
\begin{minipage}[t]{0.8\columnwidth}%
... <user> So, what do you do for work?
\end{minipage}
\\
\midrule

$R_P$ & 
\begin{minipage}[t]{0.8\columnwidth}%
<agent> I'm a stock broker.
\end{minipage}
\\
\midrule

$R_G$  & 
\begin{minipage}[t]{0.8\columnwidth}%
<agent> I'm a conversational AI agent, assisting users like you with information and tasks.
\end{minipage}
\\

\bottomrule

\end{tabular}
\vspace{5pt}
\begin{tabular}{
    @{\hspace{0pt}}p{1.1cm}@{\hspace{5pt}}|
    @{\hspace{5pt}}l@{\hspace{5pt}}
}

\multicolumn{2}{@{\hspace{0pt}}p{\dimexpr 1.1cm + 10pt + 0.8\columnwidth\relax}@{}}{%
    \textbf{Ex. 3. (SPC)}: $R_G$ provides external knowledge of the conversational topic.
} \\

\toprule
$P$ &  
\begin{minipage}[t]{0.8\columnwidth}%
<agent\_persona> I enjoy reading mysteries...
\end{minipage}
\\
\midrule

$H$ & 
\begin{minipage}[t]{0.8\columnwidth}%
... <user> Me too! I love Agatha Christie's Poirot novels.
\end{minipage}
\\
\midrule

$R_P$ & 
\begin{minipage}[t]{0.8\columnwidth}%
<agent> Me too! They're so well-written.
\end{minipage}
\\
\midrule

$R_G$  & 
\begin{minipage}[t]{0.8\columnwidth}%
<agent> Those are classics. Have you read any of her standalone novels like "The ABC Murders" or "Five Little Pigs"?
\end{minipage}
\\

\bottomrule

\end{tabular}

\begin{tablenotes}[flushleft]
    \setlength\labelsep{0pt}
    \footnotesize
    \item[$\dagger$] The persona is withheld from the LLM for generic response augmentation. 
\end{tablenotes}
\egroup 
\end{threeparttable}


\end{table}

%% file: sections/results.tex
\section{Results}
\label{sec:results}


\input{tables/best_performance}

Table~\ref{table:performance:compare} shows the performance of  pre-fusion-based \ourmethodpre and post-fusion-based \ourmethodpost, along with the four baseline methods using the \preicl and \presft schemes, across all three tasks.
We consider \presft[Hist] as the best baseline (denoted with ``$^{B}$'') as it achieves the best performance on most metrics among the baselines, and \ourmethodpost as the best challenger model (denoted with ``$^{C}$'').
%

Overall, \ourmethod substantially improves the response generation quality and consistently achieves better performance than \presft[Hist] on most metrics across all datasets.
This shows that offloading structural reasoning to the generic responses fundamentally unlocks the potential of the smaller task models. For example, \ourmethodpost achieves a massive 22.55\% improvement in ROUGE-2 and $20.88\%$ in BLEU over \presft[Hist] on the SPC task. This confirms that once a model is freed from generating the underlying conversational structure from scratch, it can synthesize personal and contextual data with high accuracy. Significant improvements are also observed across the other tasks.
Meanwhile, \ourmethodpre also shows consistent improvements across all evaluation metrics over \presft[Hist], though the improvements are slightly lower than \ourmethodpost. 

To further verify the robustness of \ourmethod, we conduct evaluations on the canonical ConvAI2 dataset. Despite the faithfulness issues inherent in the dataset, \ourmethod consistently outperforms \presft[Hist] across all metrics. This confirms that the proposed strategy to decouple contextual  grounding and personalization provides a robust, generalizable advantage over standard monolithic training.

\subsection{Generic Responses as Grounding Signal}
To illustrate the effectiveness of the generic responses, we compare \ourmethodpre and \presft[Hist], where the only difference is the inclusion of the auxiliary generic responses in the inputs.  Across all three datasets, \ourmethodpre consistently outperforms \presft[Hist] in all metrics.
For example, \ourmethodpre achieves 12.86\% / 19.15\% / 17.15\% improvements in ROUGE-1 / ROUGE-2 / BLEU scores on SPC, and improvements of 9.57\% / 42.50\% / 21.07\% on PFG. On ED, the improvements are smaller but still consistent across all metrics. These improvements demonstrate that the generic responses are able to act as a grounding prior in generation tasks and provide critical guiding signals that help the model better capture the underlying conversational intent and discourse structures. The wider performance gaps in ROUGE-2 / BLEU than ROUGE-1 also indicate \ourmethodpre is particularly better than the baseline at modeling the longer-span dependencies (e.g., bigrams and higher-order n-grams) rather than shallow lexical matching (i.e., character- or unigram-level copying). This further demonstrates that the generic responses are especially beneficial in tasks, such as PFG, where the responses are content-rich or structurally complex. 

While \ourmethod shows overall improvements across different tasks, the effectiveness can vary significantly across tasks and metrics. 
For instance, the \ourmethodpost improves 47.50\% on ROUGE-2 over \presft[Hist] on the PFG task, while the improvement is 36.67\% and 22.55\% for ED and SPC, respectively. Different degrees of improvements can also be observed on BLEU and BERT-F1 for these three tasks. 
Such variation in effectiveness reflects different task characteristics. 
For SPC, where chit-chat involves substantial explicit persona information in the responses, the grounding signals are highly beneficial when responses require both contextual relevance and persona consistency, resulting in higher improvements on BERT-F1.
For PFG, where persona usage is more implicit and responses focus more on the structure of persuasive language, the grounding signals provide critical guidance for modeling the discourse structure of the persuasive responses, and lead to significant improvements on ROUGE-2 and BLEU scores. 
For ED, where the persona data is a simple emotion tag and the responses are typically short and emotionally aligned, the generic responses are already able to capture most of the required content, which leaves less room for improvement. 
These observations highlight that the \ourmethod methods are particularly effective for scenarios where the discourse structure and content-rich generation are highly demanded.

\subsection{Pre-Fusion and Post-Fusion Architectures}
\label{sec:results:pre_vs_post}
Here we compare the \ourmethodpre and \ourmethodpost methods to investigate how architectural design choices affect performance. While both models leverage the same inputs, \ourmethodpost encodes the persona, conversation history, and generic responses separately and leverages a style embedding vector to guide the generation. 

In Table~\ref{table:performance:compare}, \ourmethodpost outperforms \ourmethodpre in most metrics. For example, \ourmethodpost improves 13.89\% / 7.38\% in ROUGE-2 / BLEU over \ourmethodpre on ED dataset and 3.51\% / 12.39\% on PFG dataset, compared to smaller improvements of 2.86\% / 3.18\% on the SPC dataset. This indicates that explicitly disentangling different input sources in post-fusion architecture leads to better understanding and utilization of the grounding signals. In contrast, the pre-fusion architecture might introduce cross-inference complexity among the three input sources, which makes it less effective than the post-fusion architecture in leveraging the auxiliary generic responses.
The post-fusion architecture is particularly beneficial for tasks like PFG and ED, where personalization is implicit or indirect (e.g., via persuasion strategies or emotional cues). It still provides marginal benefits on tasks like SPC where personalization is more explicit (e.g., directly copying from persona profile texts). 

For information compression, while both \ourmethod methods show very strong BPC performance, \ourmethodpre achieves slightly higher efficiency than \ourmethodpost.
This indicates that post-fusion-based \ourmethodpost assigns slightly lower probability to the predicted tokens than pre-fusion-based \ourmethodpre, potentially due to the implicit / indirect personalization association between persona data and the personalized responses after the style embedding layer in \ourmethodpost. Meanwhile, \ourmethodpre benefits more from the direct association in the concatenated input, especially on datasets like SPC with explicit persona-response connections. However, the higher performance on the other metrics, such as ROUGE and BLEU scores, actually indicates the post-fusion architecture of \ourmethodpost is able to generate more diverse and content-rich outputs by exploring low-probability phrases instead of repeating safe high-probability phrases.

\subsection{Comparison with In-Context Learning}
\label{sec:results:sft_vs_icl}
We further analyze the performance of the ICL methods. Notably, \preicl[Hist] achieved high Char-F1 scores on two out of three tasks, whereas standard SFT models, including the \presft and the \ourmethod variants, outperform the \preicl[Hist] models on the other metrics, particularly ROUGE, BLEU and BERT scores. 
The combination of higher Char-F1 and lower ROUGE / BLEU scores for \preicl models indicates that generated responses have more character-level matches but fewer word- or phrase-level matches, which is typically a sign that the model generates longer responses that are paraphrases of the reference responses. 
This discrepancy arises from the nature of ICL-based generation, where the pre-trained LLMs tend to generate longer, fluent but lexically diverse responses that may not match the reference responses, but still share significant character-level overlaps. 
In contrast, the responses generated by the SFT-based models are optimized based on the training data distribution, and therefore are able to generate more precise and structured responses that are more closely aligned with the reference responses.
Table~\ref{table:inference_example} shows an example of \preicl[Hist] and \ourmethodpost responses compared with the reference response from the PFG dataset. 

The results show that ICL-based approaches are better at capturing semantic adequacy and generalization, and the SFT-based models, especially our methods in the \ourmethod framework, significantly improve grounded and task-specific generation quality. Therefore, although \ourmethod methods may not achieve the highest Char-F1 scores, they still provide stronger performance in terms of content fidelity and discourse structures. 

\input{tables/example_icl_vs_sft_inference}


\subsection{Grounding Bottleneck vs. Personalization}
\label{sec:results:grounding_vs_personalization}

From Table~\ref{table:performance:compare}, the two \ourmethod methods achieve substantial improvements on the alignment-based metrics such as ROUGE and BLEU, while the improvements in semantic similarity (BERT-F1) are comparatively modest ($1\sim4\%$) across all datasets. In our human and LLM evaluations of personalization (discussed in Section~\ref{sec:results:human-llm-eval}), most outputs from \ourmethod and the baseline \presft[Hist] are rated as "similar" (approximately 60\%) with a smaller but consistent preference towards \ourmethod. 
These results suggest that the primary benefit of \ourmethod lies in better grounding and coherence in the generated responses, rather than producing substantial changes in personalization. The auxiliary generic responses play an important role as a ``content scaffold'' that addresses the grounding bottleneck of existing conversational models and guides the model towards more structured and relevant responses, while the persona data provide finer-grained stylistic adjustments.

\subsection{Human and LLM-as-a-Judge Evaluations}
\label{sec:results:human-llm-eval}

In addition to the automated evaluation metrics reported in Table~\ref{table:performance:compare}, we conduct human evaluation to assess the personalization quality of the responses. Here the evaluation is based on the test set of the SPC task, as it is easier for general annotators to judge based on the explicit persona content, compared to the implicit personalization in ED and PFG tasks.
We co-design the human evaluation with two behavioral scientists from Lirio, Inc. (\url{http://lirio.com}), including the interface, instructions, and the four preference-based evaluation questions that cover different aspects of personalization, focusing on improving clarity on the task and minimize ambiguity or misunderstandings to the annotators. 
The four questions are listed in Table~\ref{table:human_eval_question}.
The annotators consist of 13 unpaid volunteers who are independent from this research. 
The annotators are presented the conversational context (including persona, conversation history and generic response), the two candidate responses for comparison (response generated by baseline \presft[Hist], $R^{B}$, and by \ourmethodpost, $R^{C}$) along with the four questions for annotation.
Responses $R^{B}$ and $R^{C}$ are presented as ``Response A'' and ``Response B'' without revealing the source models or methods to the annotators. For each question, the annotators are asked to select one of three options (``Response A is better'', ``Response B is better'', or ``Similar'') along with a confidence rating for their answer on a scale from 1 (not confident) to 5 (very confident). Each conversation turn is rated by three different human annotators, and the responses for each question are aggregated by majority vote. Conversation turns are excluded from analysis when the three annotators all disagree. 
We have received a total of 435 annotations (145 turns; each annotated by three annotators). After filtering based on confidence level and majority vote, we retained $>$95\% of annotated turns for each question (reported in Table~\ref{table:human_eval_question} as ``\# Valid Annotations'').

\input{tables/human_eval_question}

\input{figures/human_eval_bar_chart}

The results show that the human annotators prefer \ourmethodpost more often than the baseline \presft[Hist], but mostly the gain is modest. Specifically, 15.2\% vs 26.1\% of preferred responses generated by $R^{B}$ vs $R^{C}$ for Q1, and 11.2\% vs 17.5\%, 5.6\% vs 14.8\% and  14.4\% vs 30.2\% for Q2, Q3 and Q4. 
Although a large portion of the response pairs are judged as similar, this preference indicates that our method \ourmethodpost still provides small improvements in personalization, while its primary benefits lie in leveraging generic responses to improve content richness and discourse structure. This is also consistent with the discussion in Section~\ref{sec:results:grounding_vs_personalization}.

While our human evaluation is limited to 145 annotations due to resource and time constraints, we address this limitation by supplementing the evaluations with an LLM as a judge.
%
We first evaluated the same 145 turns by using Qwen2.5-32B-Instruct model from Hugging Face~\footnote{\url{https://huggingface.co/Qwen/Qwen2.5-32B-Instruct}} as the judge. After pruning the response to match the valid annotations per question, the overall trends are consistent with human annotations. However, we observe that the LLM tends to provide more decisive preferences (rather than ``Similar'') than human annotators, mainly on Q2 and Q3. This discrepancy primarily suggests that the LLM may be more sensitive to subtle differences, whereas human annotators often consider such differences as negligible. 
Despite this discrepancy, the combined evaluation can be viewed as a robust and scalable approximation of human judgment.
We then expand the LLM-based evaluation to 2,000 turns to obtain a more representative distribution of the preferences. The result also shows an overall consistent trend that most response pairs are judged as similar, with a slightly higher preference of $R^{C}$ over $R^{B}$. 
Figure~\ref{fig:human_llm_eval} shows the results of the human evaluation, LLM evaluation on the 145 human annotated examples (denoted as ``LLM-145'') and the full 2000 annotated examples (denoted as ``LLM-2000'').

%% file: tables/best_performance.tex
\begin{table*}[t!]

\centering
\caption{Performance Comparison of \ourmethodpre and \ourmethodpost over Baseline Methods}
\label{table:performance:compare}
\begin{threeparttable}
\bgroup
\def\arraystretch{1}%
\begin{tabular}{
  @{\hspace{5pt}}l@{\hspace{5pt}}
  @{\hspace{10pt}}l@{\hspace{5pt}}
  @{\hspace{5pt}}r@{\hspace{5pt}}
  @{\hspace{0pt}}r@{\hspace{5pt}}
  @{\hspace{0pt}}r@{\hspace{5pt}}
  @{\hspace{0pt}}r@{\hspace{5pt}}
  @{\hspace{0pt}}r@{\hspace{5pt}}
  @{\hspace{0pt}}r@{\hspace{5pt}}
  @{\hspace{0pt}}r@{\hspace{5pt}}
  @{\hspace{0pt}}r@{\hspace{5pt}}
  @{\hspace{0pt}}r@{\hspace{5pt}}
}
\toprule
                    Task & Method  & Char-F1 & ROUGE-1 & ROUGE-2 & ROUGE-L & BLEU & BPC$^{\downarrow}$ & BERT-P & BERT-R & BERT-F1 \\

\midrule

\multirow{6}{30pt}{ED}      & \preicl[Hist]                &  \bf{0.168} &  0.142 &  0.019 &  0.117 &  2.216 &  2.350 &  \bf{0.587} &  0.510 &  0.544  \\
                            & \preicl[TST]                 &  0.155 &  0.125 &  0.014 &  0.105 &  2.353 &  2.790 &  0.569 &  0.510 &  0.536 \\
                            
                            & \presft[Hist]$^{B}$          &  0.132 &  0.165 &  0.030 &  0.148 &  3.595 &  1.030 &  0.556 &  0.578 &  0.564  \\
                            & \presft[TST]                 &  0.127 &  0.158 &  0.029 &  0.142 &  3.346 &  1.082 &  0.548 &  0.570 &  0.556  \\
                            & \ourmethodpre                &  0.138 &  0.173 &  0.036 &  0.156 &  3.806 &  \bf{1.022} &  0.561 &  0.581 &  0.568  \\
                            & \ourmethodpost$^{C}$         &  0.146 &  \bf{0.182} &  \bf{0.041} &  \bf{0.164} &  \bf{4.087} &  1.023 &  0.567 &  \bf{0.584} &  \bf{0.573}  \\
                            
                            \cline{2-11}
                            & imprv (\%) \ourmethodpre     &   4.55 &   4.85 &  20.00 &   5.41 &   5.87 &   0.78 &   0.90 &   0.52 &   0.71  \\
                            & imprv (\%) \ourmethodpost    &  10.61 &  10.30 &  36.67 &  10.81 &  13.69 &   0.68 &   1.98 &   1.04 &   1.60  \\

\midrule
\multirow{6}{30pt}{PFG}     & \preicl[Hist]                &  \bf{0.191} &  0.167 &  0.028 &  0.126 &  2.302 &  1.736 &  \bf{0.588} &  0.554 &  0.568  \\
                            & \preicl[TST]                 &  0.190 &  0.167 &  0.026 &  0.122 &  2.111 &  2.227 &  0.581 &  0.538 &  0.556  \\
                            
                            & \presft[Hist]$^{B}$          &  0.166 &  0.188 &  0.040 &  0.149 &  3.394 &  \bf{0.923} &  0.553 &  0.581 &  0.564  \\
                            & \presft[TST]                 &  0.168 &  0.186 &  0.046 &  0.153 &  3.368 &  0.964 &  0.552 &  0.573 &  0.559  \\
                            & \ourmethodpre                &  0.182 &  0.206 &  0.057 &  0.167 &  4.109 &  \bf{0.923} &  0.563 &  0.585 &  0.571  \\
                            & \ourmethodpost$^{C}$         &  0.186 &  \bf{0.214} &  \bf{0.059} &  \bf{0.175} &  \bf{4.618} &  0.938 &  0.569 &  \bf{0.587} &  \bf{0.575}  \\
                            
                            \cline{2-11}
                            & imprv (\%) \ourmethodpre     &   9.64 &   9.57 &  42.50 &  12.08 &  21.07 &   0.00 &   1.81 &   0.69 &   1.24  \\
                            & imprv (\%) \ourmethodpost    &  12.05 &  13.83 &  47.50 &  17.45 &  36.06 &  -1.63 &   2.89 &   1.03 &   1.95  \\

\midrule
\multirow{6}{30pt}{SPC}     & \preicl[Hist]                &  0.231 &  0.231 &  0.088 &  0.207 &  5.967 &  1.333 &  0.674 &  0.583 &  0.622  \\
                            & \preicl[TST]                 &  0.167 &  0.151 &  0.033 &  0.135 &  3.237 &  2.127 &  0.625 &  0.534 &  0.573  \\
                            
                            & \presft[Hist]$^{B}$          &  0.317 &  0.381 &  0.235 &  0.365 & 17.444 &  0.384 &  0.709 &  0.704 &  0.704  \\
                            & \presft[TST]                 &  0.285 &  0.348 &  0.200 &  0.333 & 14.500 &  0.466 &  0.694 &  0.690 &  0.689  \\
                            & \ourmethodpre                &  0.358 &  0.430 &  0.280 &  0.414 & 20.436 &  \bf{0.381} &  0.731 &  0.732 &  0.729  \\
                            & \ourmethodpost$^{C}$         &  \bf{0.366} &  \bf{0.438} &  \bf{0.288} &  \bf{0.423} & \bf{21.086} &  0.403 &  \bf{0.741} &  \bf{0.734} &  \bf{0.735}  \\
                            
                            \cline{2-11}
                            & imprv (\%) \ourmethodpre     &  12.93 &  12.86 &  19.15 &  13.42 &  17.15 &   0.78 &   3.10 &   3.98 &   3.55  \\
                            & imprv (\%) \ourmethodpost    &  15.46 &  14.96 &  22.55 &  15.89 &  20.88 &  -4.95 &   4.51 &   4.26 &   4.40  \\

                            
                            

\bottomrule
\end{tabular}

\begin{tablenotes}[flushleft]
    \setlength\labelsep{0pt}
    \footnotesize
    \item For each task, values in \textbf{bold} represent the best performance of the corresponding metric for each task. Models denoted with ``$\cdot$ $^{B}$'' (baseline) and ``$\cdot$ $^{C}$'' (challenger) are the best baseline and \ourmethod models, respectively. ``imprv (\%)'' measures the percentage improvement of the \ourmethod models over the ``$\cdot$ $^{B}$'' models.

\end{tablenotes}
\egroup 
\end{threeparttable}
\end{table*}

%% file: tables/example_icl_vs_sft_inference.tex
\begin{table}
\centering
\footnotesize

\caption{Example of \preicl[Hist] and \ourmethodpost Responses - PFG}
\vspace{-5pt}
\label{table:inference_example}

\begin{threeparttable}
\bgroup
\def\arraystretch{1}%
\begin{tabular}{
    @{\hspace{0pt}}p{1.1cm}@{\hspace{5pt}}|
    @{\hspace{5pt}}l@{\hspace{5pt}}
}

\multicolumn{2}{@{\hspace{0pt}}p{\dimexpr 1.1cm + 10pt + 0.8\columnwidth\relax}@{}}{%
    \preicl[Hist] tend to generate lengthy and rephrased responses with more character-level matches but fewer word- and phrase-level matches, whereas the  \ourmethodpost responses are precise and hit the core semantics (indicated by \underline{underlined words}). 
} \\

\toprule

Reference Response  & 
\begin{minipage}[t]{0.8\columnwidth}%
<agent> This \underline{organization} is called \underline{Save the Children} and your donation can help ensure \underline{children's rights} to health, education, and safety.
\end{minipage}
\\
\midrule

\preicl[Hist] Response  & 
\begin{minipage}[t]{0.8\columnwidth}%
<agent> That's great question, let me tell you more about \underline{Save the Children}, it's a non-profit organization that works tirelessly to provide aid to children in need worldwide, focusing on \underline{education}, \underline{health}, and emergency relief. They have programs in over 120 countries, including the United States, and have helped countless children overcome adversity. Would you consider learning more about their mission and how your donation could make a tangible impact?
\end{minipage}
\\

\midrule
\ourmethodpost Response  & 
\begin{minipage}[t]{0.8\columnwidth}%
<agent> \underline{Save the Children} is an international non-governmental \underline{organization} that promotes \underline{children's rights}, provides relief and helps support children in developing countries.
\end{minipage}
\\

\bottomrule
\end{tabular}

\egroup 
\end{threeparttable}
\vspace{-5pt}

\end{table}

%% file: tables/human_eval_question.tex
\begin{table}
\centering
\footnotesize
\caption{Human Evaluation Questions}
\label{table:human_eval_question}

\begin{threeparttable}
\bgroup
\def\arraystretch{1}%
\begin{tabular}{
    @{\hspace{5pt}}p{1cm}@{\hspace{5pt}}
    @{\hspace{5pt}}p{6.0cm}@{\hspace{5pt}}
    @{\hspace{5pt}}r@{\hspace{5pt}}
}
\toprule

Category & Evaluation Question & \#V.A.\\

\midrule

Q1 \par Naturalness 
    & Which response makes more sense (grammatically correct, fluent, and coherent) in the context of the conversation?
    & 138
\\
\midrule

Q2 \par Accuracy 
    & Which response is better at correctly reflecting the facts stated in the provided persona without adding contradictory or unmentioned traits?
    & 143
\\
\midrule

Q3 \par Value 
    & Which response adds more valuable personalization beyond what the Generic Response already provides?
    & 142
\\
\midrule

Q4 \par Balance 
    & Which response does better at incorporating the persona without sacrificing the conversation quality in terms of helpfulness, engagement, relevance, or appropriateness?
    & 139
\\

\bottomrule
\end{tabular}

\begin{tablenotes}[flushleft]
    \setlength\labelsep{0pt}
    \footnotesize
    \item 
    ``\#V.A.'' column has the number of examples with valid annotations (out of the 145 annotated examples) after majority voting of the three annotators for each question. 
\end{tablenotes}
\vspace{-10pt}
\egroup 
\end{threeparttable}

\end{table}

%% file: figures/human_eval_bar_chart.tex
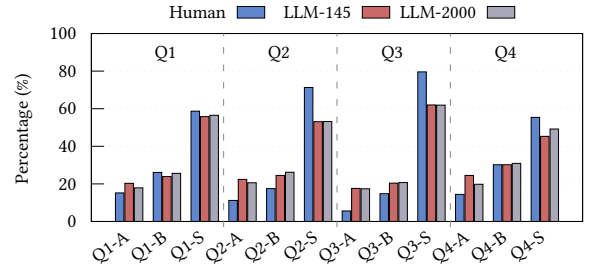
\begin{figure}[t]
    \centering
    \footnotesize

    \input{annotation_plot.tex}
    \vspace{-20pt}
    \caption{Human and LLM annotation results.
      Each group of three bars corresponds to one choice of one question.
      The choices (A / B / S) correspond to annotations that $R^{B}$ is better, $R^{C}$ is better or similar, respectively. 
      Within each group the three colors indicate annotator types
      (Human, LLM-145, LLM-2000), demonstrating cross-annotator consistency.
    }
    \vspace{-10pt}
    \label{fig:human_llm_eval}
\end{figure}

%% file: annotation_plot.tex
\begingroup
  \makeatletter
  \providecommand\color[2][]{%
    \GenericError{(gnuplot) \space\space\space\@spaces}{%
      Package color not loaded in conjunction with
      terminal option `colourtext'%
    }{See the gnuplot documentation for explanation.%
    }{Either use 'blacktext' in gnuplot or load the package
      color.sty in LaTeX.}%
    \renewcommand\color[2][]{}%
  }%
  \providecommand\includegraphics[2][]{%
    \GenericError{(gnuplot) \space\space\space\@spaces}{%
      Package graphicx or graphics not loaded%
    }{See the gnuplot documentation for explanation.%
    }{The gnuplot epslatex terminal needs graphicx.sty or graphics.sty.}%
    \renewcommand\includegraphics[2][]{}%
  }%
  \providecommand\rotatebox[2]{#2}%
  \@ifundefined{ifGPcolor}{%
    \newif\ifGPcolor
    \GPcolortrue
  }{}%
  \@ifundefined{ifGPblacktext}{%
    \newif\ifGPblacktext
    \GPblacktexttrue
  }{}%
  \let\gplgaddtomacro\g@addto@macro
  \gdef\gplbacktext{}%
  \gdef\gplfronttext{}%
  \makeatother
  \ifGPblacktext
    \def\colorrgb#1{}%
    \def\colorgray#1{}%
  \else
    \ifGPcolor
      \def\colorrgb#1{\color[rgb]{#1}}%
      \def\colorgray#1{\color[gray]{#1}}%
      \expandafter\def\csname LTw\endcsname{\color{white}}%
      \expandafter\def\csname LTb\endcsname{\color{black}}%
      \expandafter\def\csname LTa\endcsname{\color{black}}%
      \expandafter\def\csname LT0\endcsname{\color[rgb]{1,0,0}}%
      \expandafter\def\csname LT1\endcsname{\color[rgb]{0,1,0}}%
      \expandafter\def\csname LT2\endcsname{\color[rgb]{0,0,1}}%
      \expandafter\def\csname LT3\endcsname{\color[rgb]{1,0,1}}%
      \expandafter\def\csname LT4\endcsname{\color[rgb]{0,1,1}}%
      \expandafter\def\csname LT5\endcsname{\color[rgb]{1,1,0}}%
      \expandafter\def\csname LT6\endcsname{\color[rgb]{0,0,0}}%
      \expandafter\def\csname LT7\endcsname{\color[rgb]{1,0.3,0}}%
      \expandafter\def\csname LT8\endcsname{\color[rgb]{0.5,0.5,0.5}}%
    \else
      \def\colorrgb#1{\color{black}}%
      \def\colorgray#1{\color[gray]{#1}}%
      \expandafter\def\csname LTw\endcsname{\color{white}}%
      \expandafter\def\csname LTb\endcsname{\color{black}}%
      \expandafter\def\csname LTa\endcsname{\color{black}}%
      \expandafter\def\csname LT0\endcsname{\color{black}}%
      \expandafter\def\csname LT1\endcsname{\color{black}}%
      \expandafter\def\csname LT2\endcsname{\color{black}}%
      \expandafter\def\csname LT3\endcsname{\color{black}}%
      \expandafter\def\csname LT4\endcsname{\color{black}}%
      \expandafter\def\csname LT5\endcsname{\color{black}}%
      \expandafter\def\csname LT6\endcsname{\color{black}}%
      \expandafter\def\csname LT7\endcsname{\color{black}}%
      \expandafter\def\csname LT8\endcsname{\color{black}}%
    \fi
  \fi
    \setlength{\unitlength}{0.0500bp}%
    \ifx\gptboxheight\undefined%
      \newlength{\gptboxheight}%
      \newlength{\gptboxwidth}%
      \newsavebox{\gptboxtext}%
    \fi%
    \setlength{\fboxrule}{0.5pt}%
    \setlength{\fboxsep}{1pt}%
    \definecolor{tbcol}{rgb}{1,1,1}%
\begin{picture}(4800.00,2260.00)%
    \gplgaddtomacro\gplbacktext{%
      \csname LTb\endcsname
      \put(692,470){\makebox(0,0)[r]{\strut{}$0$}}%
      \csname LTb\endcsname
      \put(692,752){\makebox(0,0)[r]{\strut{}$20$}}%
      \csname LTb\endcsname
      \put(692,1034){\makebox(0,0)[r]{\strut{}$40$}}%
      \csname LTb\endcsname
      \put(692,1316){\makebox(0,0)[r]{\strut{}$60$}}%
      \csname LTb\endcsname
      \put(692,1598){\makebox(0,0)[r]{\strut{}$80$}}%
      \csname LTb\endcsname
      \put(692,1880){\makebox(0,0)[r]{\strut{}$100$}}%
      \csname LTb\endcsname
      \put(1076,370){\rotatebox{35.00}{\makebox(0,0)[r]{\strut{}Q1-A}}}%
      \csname LTb\endcsname
      \put(1360,370){\rotatebox{35.00}{\makebox(0,0)[r]{\strut{}Q1-B}}}%
      \csname LTb\endcsname
      \put(1643,370){\rotatebox{35.00}{\makebox(0,0)[r]{\strut{}Q1-S}}}%
      \csname LTb\endcsname
      \put(1927,370){\rotatebox{35.00}{\makebox(0,0)[r]{\strut{}Q2-A}}}%
      \csname LTb\endcsname
      \put(2210,370){\rotatebox{35.00}{\makebox(0,0)[r]{\strut{}Q2-B}}}%
      \csname LTb\endcsname
      \put(2493,370){\rotatebox{35.00}{\makebox(0,0)[r]{\strut{}Q2-S}}}%
      \csname LTb\endcsname
      \put(2777,370){\rotatebox{35.00}{\makebox(0,0)[r]{\strut{}Q3-A}}}%
      \csname LTb\endcsname
      \put(3060,370){\rotatebox{35.00}{\makebox(0,0)[r]{\strut{}Q3-B}}}%
      \csname LTb\endcsname
      \put(3344,370){\rotatebox{35.00}{\makebox(0,0)[r]{\strut{}Q3-S}}}%
      \csname LTb\endcsname
      \put(3627,370){\rotatebox{35.00}{\makebox(0,0)[r]{\strut{}Q4-A}}}%
      \csname LTb\endcsname
      \put(3911,370){\rotatebox{35.00}{\makebox(0,0)[r]{\strut{}Q4-B}}}%
      \csname LTb\endcsname
      \put(4194,370){\rotatebox{35.00}{\makebox(0,0)[r]{\strut{}Q4-S}}}%
      \csname LTb\endcsname
      \put(1360,1739){\makebox(0,0){\strut{}Q1}}%
      \csname LTb\endcsname
      \put(2210,1739){\makebox(0,0){\strut{}Q2}}%
      \csname LTb\endcsname
      \put(3060,1739){\makebox(0,0){\strut{}Q3}}%
      \csname LTb\endcsname
      \put(3911,1739){\makebox(0,0){\strut{}Q4}}%
    }%
    \gplgaddtomacro\gplfronttext{%
      \csname LTb\endcsname
      \put(1814,2039){\makebox(0,0)[r]{\strut{}Human}}%
      \csname LTb\endcsname
      \put(2752,2039){\makebox(0,0)[r]{\strut{}LLM-145}}%
      \csname LTb\endcsname
      \put(3690,2039){\makebox(0,0)[r]{\strut{}LLM-2000}}%
      \csname LTb\endcsname
      \put(296,1175){\rotatebox{-270.00}{\makebox(0,0){\strut{}Percentage (\%)}}}%
    }%
    \gplbacktext
    \put(0,0){\includegraphics[width={240.00bp},height={113.00bp}]{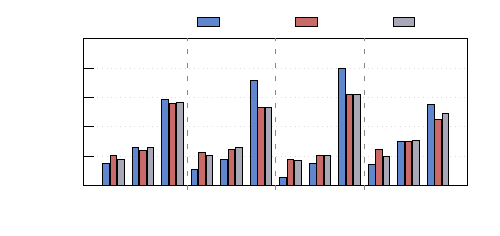}}%
    \gplfronttext
  \end{picture}%
\endgroup

%% file: sections/ablation.tex
\section{Ablation Study}
\label{sec:ablation}

\subsection{Effect of Contrastive JEPA Learning}
\label{sec:ablation:jepa}

\input{figures/jepa_trend}

We conduct experiments with different C-JEPA loss weights $\beta$ on both \ourmethodpre and \ourmethodpost methods ($\alpha$ is set to 0). 
Intuitively, effective C-JEPA learning will encourage the context representation to be similar to the personalized responses, and therefore, will create better discourse alignment as measured by ROUGE-2 and BLEU scores.
Figure~\ref{fig:jepa_trend} shows the performance with respect to $\beta$ on the ED and SPC datasets. 

For the post-fusion-based \ourmethodpost method, the C-JEPA loss consistently improves performance on both datasets. On the ED dataset, when $\beta$=0, both ROUGE-2 and BLEU scores are low. When $\beta$ increases from zero, performance improves and peaks at $\beta$=0.5. As $\beta$ continues to increase, both ROUGE-2 and BLEU start to decrease, as the model places too much emphasis on the contrastive similarities among the responses and context, causing the model to be distracted from modeling the final response. A similar trend is observed on the SPC dataset, with an optimal $\beta$ of 0.05. 

For the pre-fusion-based \ourmethodpre method, the C-JEPA loss yields limited gains on the ED dataset and the performance does not change significantly as $\beta$ changes. For the SPC dataset, we do observe that a non-zero $\beta$ does yield better performance than $\beta$=0. However, as $\beta$ continues to increase beyond zero, the change does not make a significant difference. 

These results indicate that the post-fusion architecture of \ourmethodpost is more robust and better able to exploit the contrastive differences between the generic responses and personalized responses. In contrast, the pre-fusion architecture of \ourmethodpre entangles the embedding of different inputs in a single encoder, which potentially increased the learning complexity of the model, and thus, is less responsive to such differentiating signals, especially when the connection between the persona profile and the response is implicit, as in the ED dataset. The \ourmethodpre model only demonstrates limited responsiveness when the persona-response connection is more explicit and the discourse structures are more predictable.

\subsection{Effect of Style Removal Loss}
\label{sec:ablation:style_removal}

\input{figures/bw_trend}

To analyze the effect of the style removal loss, we experiment with different $\alpha$ values on \ourmethodpost method, with $\beta$ set to 0. Figure~\ref{fig:bw_trend} shows the performance with respect to $\alpha$ on the ED and SPC datasets. 
For the ED dataset, the performance is not impacted much by the $\alpha$ values. 
For the SPC dataset, a non-zero $\alpha$ value yields slightly higher performance compared to when $\alpha$=0. However, further increase in non-zero $\alpha$ values does not yield further improvements on the performance.
These indicate that the style removal loss and the back-translation strategy are particularly effective for tasks such as SPC, where the connection between the persona profile and the responses is explicit, allowing the style removal network to better capture and preserve response semantics. For tasks like ED where the persona-response connections are implicit, the style removal strategy is less effective.

\subsection{Response Generation vs. Text Style Transfer}
\label{sec:ablation:tst}

We provide some further analysis on the performance of TST approaches and explain why we do not adopt pure TST approach in the \ourmethod methods. Table~\ref{table:performance:compare} shows the performance of the two response generation paradigms, using a traditional generation based on conversation history (``\texttt{Hist}'') and pure style transfer based on the generic responses (``\texttt{TST}'').
These results demonstrate consistent and significant performance drops of both \preicl[TST] and \presft[TST] from \preicl[Hist] and \presft[Hist] across all three datasets. 
We conducted further experiments to adapt \ourmethod to the TST paradigm, by using $E_{R_G}$ as the encoder hidden states as described in Section~\ref{sec:method:style-decoder}. These experiments  also show a significant performance decrease compared to the original \ourmethod.

These observations reflect the fundamentally different natures of the conversation and text style transfer problems. 
In TST problems, source and target texts are expected to share the same or highly similar semantics and the model is responsible for adjusting the output in different styles or tones, whereas in conversation problems, the response generation not only relies on the semantics but also broader contextual information from the conversation history. 
While the generic responses provide valuable structural and content-rich signals to aid the model's learning, the model also loses access to critical signals from the conversation history that are unrecoverable after highly lossy compression process of generating generic responses. 
Therefore, our \ourmethod methods adopt the strategy that retains the fundamental signals from the conversation history while leveraging the values of the auxiliary generic responses.

%% file: figures/jepa_trend.tex
\begin{figure}[ht]
    \vspace{-5pt}
    \centering

    \begin{subfigure}[ht]{0.49\linewidth}
        {\tiny






        \input{gpl_jepa_trend_ed.tex}
        }
    \end{subfigure}%
    \hfill
    \begin{subfigure}[ht]{0.49\linewidth}
        {\tiny






        \input{gpl_jepa_trend_spc.tex}
        }
    \end{subfigure}
    
    \vspace{-105pt}
    \begin{tikzpicture}
        \begin{axis}[
            hide axis,
            xmin=0, xmax=1, ymin=0, ymax=1,
            legend columns=4,
            legend style={
                draw=none,
                /tikz/every even column/.append style={column sep=0.5em},
                font=\tiny,
                inner sep=1pt,
            },
        ]
        \addlegendimage{color=c1, mark=*, mark size=1pt, line width=0.8pt}
        \addlegendentry{ROUGE-2 \ourmethodpost}
        \addlegendimage{color=c2, mark=triangle*, mark size=1pt, line width=0.8pt}
        \addlegendentry{BLEU \ourmethodpost}
        \addlegendimage{color=c3, mark=square*, mark size=1pt, line width=0.8pt}
        \addlegendentry{ROUGE-2 \ourmethodpre}
        \addlegendimage{color=c4, mark=diamond*, mark size=1pt, line width=0.8pt}
        \addlegendentry{BLEU \ourmethodpre}
        \end{axis}
    \end{tikzpicture}
    \vspace{50pt}
    \caption{\ourmethod Performance w.r.t. JEPA Weight $\beta$}
    \vspace{-5pt}
    \label{fig:jepa_trend}
\end{figure}
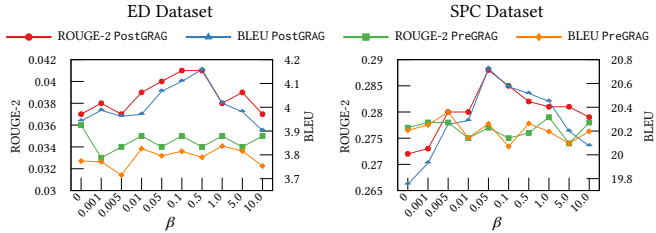

%% file: gpl_jepa_trend_ed.tex
\begingroup
  \makeatletter
  \providecommand\color[2][]{%
    \GenericError{(gnuplot) \space\space\space\@spaces}{%
      Package color not loaded in conjunction with
      terminal option `colourtext'%
    }{See the gnuplot documentation for explanation.%
    }{Either use 'blacktext' in gnuplot or load the package
      color.sty in LaTeX.}%
    \renewcommand\color[2][]{}%
  }%
  \providecommand\includegraphics[2][]{%
    \GenericError{(gnuplot) \space\space\space\@spaces}{%
      Package graphicx or graphics not loaded%
    }{See the gnuplot documentation for explanation.%
    }{The gnuplot epslatex terminal needs graphicx.sty or graphics.sty.}%
    \renewcommand\includegraphics[2][]{}%
  }%
  \providecommand\rotatebox[2]{#2}%
  \@ifundefined{ifGPcolor}{%
    \newif\ifGPcolor
    \GPcolortrue
  }{}%
  \@ifundefined{ifGPblacktext}{%
    \newif\ifGPblacktext
    \GPblacktexttrue
  }{}%
  \let\gplgaddtomacro\g@addto@macro
  \gdef\gplbacktext{}%
  \gdef\gplfronttext{}%
  \makeatother
  \ifGPblacktext
    \def\colorrgb#1{}%
    \def\colorgray#1{}%
  \else
    \ifGPcolor
      \def\colorrgb#1{\color[rgb]{#1}}%
      \def\colorgray#1{\color[gray]{#1}}%
      \expandafter\def\csname LTw\endcsname{\color{white}}%
      \expandafter\def\csname LTb\endcsname{\color{black}}%
      \expandafter\def\csname LTa\endcsname{\color{black}}%
      \expandafter\def\csname LT0\endcsname{\color[rgb]{1,0,0}}%
      \expandafter\def\csname LT1\endcsname{\color[rgb]{0,1,0}}%
      \expandafter\def\csname LT2\endcsname{\color[rgb]{0,0,1}}%
      \expandafter\def\csname LT3\endcsname{\color[rgb]{1,0,1}}%
      \expandafter\def\csname LT4\endcsname{\color[rgb]{0,1,1}}%
      \expandafter\def\csname LT5\endcsname{\color[rgb]{1,1,0}}%
      \expandafter\def\csname LT6\endcsname{\color[rgb]{0,0,0}}%
      \expandafter\def\csname LT7\endcsname{\color[rgb]{1,0.3,0}}%
      \expandafter\def\csname LT8\endcsname{\color[rgb]{0.5,0.5,0.5}}%
    \else
      \def\colorrgb#1{\color{black}}%
      \def\colorgray#1{\color[gray]{#1}}%
      \expandafter\def\csname LTw\endcsname{\color{white}}%
      \expandafter\def\csname LTb\endcsname{\color{black}}%
      \expandafter\def\csname LTa\endcsname{\color{black}}%
      \expandafter\def\csname LT0\endcsname{\color{black}}%
      \expandafter\def\csname LT1\endcsname{\color{black}}%
      \expandafter\def\csname LT2\endcsname{\color{black}}%
      \expandafter\def\csname LT3\endcsname{\color{black}}%
      \expandafter\def\csname LT4\endcsname{\color{black}}%
      \expandafter\def\csname LT5\endcsname{\color{black}}%
      \expandafter\def\csname LT6\endcsname{\color{black}}%
      \expandafter\def\csname LT7\endcsname{\color{black}}%
      \expandafter\def\csname LT8\endcsname{\color{black}}%
    \fi
  \fi
    \setlength{\unitlength}{0.0500bp}%
    \ifx\gptboxheight\undefined%
      \newlength{\gptboxheight}%
      \newlength{\gptboxwidth}%
      \newsavebox{\gptboxtext}%
    \fi%
    \setlength{\fboxrule}{0.5pt}%
    \setlength{\fboxsep}{1pt}%
    \definecolor{tbcol}{rgb}{1,1,1}%
\begin{picture}(2540.00,1980.00)%
    \gplgaddtomacro\gplbacktext{%
      \csname LTb\endcsname
      \put(403,588){\makebox(0,0)[r]{\strut{}0.03}}%
      \csname LTb\endcsname
      \put(403,751){\makebox(0,0)[r]{\strut{}0.032}}%
      \csname LTb\endcsname
      \put(403,914){\makebox(0,0)[r]{\strut{}0.034}}%
      \csname LTb\endcsname
      \put(403,1078){\makebox(0,0)[r]{\strut{}0.036}}%
      \csname LTb\endcsname
      \put(403,1241){\makebox(0,0)[r]{\strut{}0.038}}%
      \csname LTb\endcsname
      \put(403,1404){\makebox(0,0)[r]{\strut{}0.04}}%
      \csname LTb\endcsname
      \put(403,1567){\makebox(0,0)[r]{\strut{}0.042}}%
      \csname LTb\endcsname
      \put(579,559){\rotatebox{45.00}{\makebox(0,0)[r]{\strut{}0}}}%
      \csname LTb\endcsname
      \put(730,559){\rotatebox{45.00}{\makebox(0,0)[r]{\strut{}0.001}}}%
      \csname LTb\endcsname
      \put(882,559){\rotatebox{45.00}{\makebox(0,0)[r]{\strut{}0.005}}}%
      \csname LTb\endcsname
      \put(1033,559){\rotatebox{45.00}{\makebox(0,0)[r]{\strut{}0.01}}}%
      \csname LTb\endcsname
      \put(1184,559){\rotatebox{45.00}{\makebox(0,0)[r]{\strut{}0.05}}}%
      \csname LTb\endcsname
      \put(1335,559){\rotatebox{45.00}{\makebox(0,0)[r]{\strut{}0.1}}}%
      \csname LTb\endcsname
      \put(1486,559){\rotatebox{45.00}{\makebox(0,0)[r]{\strut{}0.5}}}%
      \csname LTb\endcsname
      \put(1637,559){\rotatebox{45.00}{\makebox(0,0)[r]{\strut{}1.0}}}%
      \csname LTb\endcsname
      \put(1789,559){\rotatebox{45.00}{\makebox(0,0)[r]{\strut{}5.0}}}%
      \csname LTb\endcsname
      \put(1940,559){\rotatebox{45.00}{\makebox(0,0)[r]{\strut{}10.0}}}%
      \csname LTb\endcsname
      \put(2116,677){\makebox(0,0)[l]{\strut{}3.7}}%
      \csname LTb\endcsname
      \put(2116,855){\makebox(0,0)[l]{\strut{}3.8}}%
      \csname LTb\endcsname
      \put(2116,1033){\makebox(0,0)[l]{\strut{}3.9}}%
      \csname LTb\endcsname
      \put(2116,1211){\makebox(0,0)[l]{\strut{}4}}%
      \csname LTb\endcsname
      \put(2116,1389){\makebox(0,0)[l]{\strut{}4.1}}%
      \csname LTb\endcsname
      \put(2116,1567){\makebox(0,0)[l]{\strut{}4.2}}%
    }%
    \gplgaddtomacro\gplfronttext{%
      \csname LTb\endcsname
      \put(107,1077){\rotatebox{-270.00}{\makebox(0,0){\strut{}ROUGE-2}}}%
      \csname LTb\endcsname
      \put(2306,1077){\rotatebox{-270.00}{\makebox(0,0){\strut{}BLEU}}}%
      \csname LTb\endcsname
      \put(1259,351){\makebox(0,0){\strut{}$\beta$}}%
      \csname LTb\endcsname
      \put(1259,1927){\makebox(0,0){\strut{}\footnotesize ED Dataset}}%
    }%
    \gplbacktext
    \put(0,0){\includegraphics[width={127.00bp},height={99.00bp}]{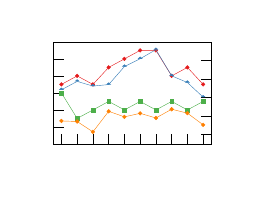}}%
    \gplfronttext
  \end{picture}%
\endgroup

%% file: gpl_jepa_trend_spc.tex
\begingroup
  \makeatletter
  \providecommand\color[2][]{%
    \GenericError{(gnuplot) \space\space\space\@spaces}{%
      Package color not loaded in conjunction with
      terminal option `colourtext'%
    }{See the gnuplot documentation for explanation.%
    }{Either use 'blacktext' in gnuplot or load the package
      color.sty in LaTeX.}%
    \renewcommand\color[2][]{}%
  }%
  \providecommand\includegraphics[2][]{%
    \GenericError{(gnuplot) \space\space\space\@spaces}{%
      Package graphicx or graphics not loaded%
    }{See the gnuplot documentation for explanation.%
    }{The gnuplot epslatex terminal needs graphicx.sty or graphics.sty.}%
    \renewcommand\includegraphics[2][]{}%
  }%
  \providecommand\rotatebox[2]{#2}%
  \@ifundefined{ifGPcolor}{%
    \newif\ifGPcolor
    \GPcolortrue
  }{}%
  \@ifundefined{ifGPblacktext}{%
    \newif\ifGPblacktext
    \GPblacktexttrue
  }{}%
  \let\gplgaddtomacro\g@addto@macro
  \gdef\gplbacktext{}%
  \gdef\gplfronttext{}%
  \makeatother
  \ifGPblacktext
    \def\colorrgb#1{}%
    \def\colorgray#1{}%
  \else
    \ifGPcolor
      \def\colorrgb#1{\color[rgb]{#1}}%
      \def\colorgray#1{\color[gray]{#1}}%
      \expandafter\def\csname LTw\endcsname{\color{white}}%
      \expandafter\def\csname LTb\endcsname{\color{black}}%
      \expandafter\def\csname LTa\endcsname{\color{black}}%
      \expandafter\def\csname LT0\endcsname{\color[rgb]{1,0,0}}%
      \expandafter\def\csname LT1\endcsname{\color[rgb]{0,1,0}}%
      \expandafter\def\csname LT2\endcsname{\color[rgb]{0,0,1}}%
      \expandafter\def\csname LT3\endcsname{\color[rgb]{1,0,1}}%
      \expandafter\def\csname LT4\endcsname{\color[rgb]{0,1,1}}%
      \expandafter\def\csname LT5\endcsname{\color[rgb]{1,1,0}}%
      \expandafter\def\csname LT6\endcsname{\color[rgb]{0,0,0}}%
      \expandafter\def\csname LT7\endcsname{\color[rgb]{1,0.3,0}}%
      \expandafter\def\csname LT8\endcsname{\color[rgb]{0.5,0.5,0.5}}%
    \else
      \def\colorrgb#1{\color{black}}%
      \def\colorgray#1{\color[gray]{#1}}%
      \expandafter\def\csname LTw\endcsname{\color{white}}%
      \expandafter\def\csname LTb\endcsname{\color{black}}%
      \expandafter\def\csname LTa\endcsname{\color{black}}%
      \expandafter\def\csname LT0\endcsname{\color{black}}%
      \expandafter\def\csname LT1\endcsname{\color{black}}%
      \expandafter\def\csname LT2\endcsname{\color{black}}%
      \expandafter\def\csname LT3\endcsname{\color{black}}%
      \expandafter\def\csname LT4\endcsname{\color{black}}%
      \expandafter\def\csname LT5\endcsname{\color{black}}%
      \expandafter\def\csname LT6\endcsname{\color{black}}%
      \expandafter\def\csname LT7\endcsname{\color{black}}%
      \expandafter\def\csname LT8\endcsname{\color{black}}%
    \fi
  \fi
    \setlength{\unitlength}{0.0500bp}%
    \ifx\gptboxheight\undefined%
      \newlength{\gptboxheight}%
      \newlength{\gptboxwidth}%
      \newsavebox{\gptboxtext}%
    \fi%
    \setlength{\fboxrule}{0.5pt}%
    \setlength{\fboxsep}{1pt}%
    \definecolor{tbcol}{rgb}{1,1,1}%
\begin{picture}(2540.00,1980.00)%
    \gplgaddtomacro\gplbacktext{%
      \csname LTb\endcsname
      \put(403,588){\makebox(0,0)[r]{\strut{}0.265}}%
      \csname LTb\endcsname
      \put(403,784){\makebox(0,0)[r]{\strut{}0.27}}%
      \csname LTb\endcsname
      \put(403,980){\makebox(0,0)[r]{\strut{}0.275}}%
      \csname LTb\endcsname
      \put(403,1175){\makebox(0,0)[r]{\strut{}0.28}}%
      \csname LTb\endcsname
      \put(403,1371){\makebox(0,0)[r]{\strut{}0.285}}%
      \csname LTb\endcsname
      \put(403,1567){\makebox(0,0)[r]{\strut{}0.29}}%
      \csname LTb\endcsname
      \put(579,559){\rotatebox{45.00}{\makebox(0,0)[r]{\strut{}0}}}%
      \csname LTb\endcsname
      \put(730,559){\rotatebox{45.00}{\makebox(0,0)[r]{\strut{}0.001}}}%
      \csname LTb\endcsname
      \put(882,559){\rotatebox{45.00}{\makebox(0,0)[r]{\strut{}0.005}}}%
      \csname LTb\endcsname
      \put(1033,559){\rotatebox{45.00}{\makebox(0,0)[r]{\strut{}0.01}}}%
      \csname LTb\endcsname
      \put(1184,559){\rotatebox{45.00}{\makebox(0,0)[r]{\strut{}0.05}}}%
      \csname LTb\endcsname
      \put(1335,559){\rotatebox{45.00}{\makebox(0,0)[r]{\strut{}0.1}}}%
      \csname LTb\endcsname
      \put(1486,559){\rotatebox{45.00}{\makebox(0,0)[r]{\strut{}0.5}}}%
      \csname LTb\endcsname
      \put(1637,559){\rotatebox{45.00}{\makebox(0,0)[r]{\strut{}1.0}}}%
      \csname LTb\endcsname
      \put(1789,559){\rotatebox{45.00}{\makebox(0,0)[r]{\strut{}5.0}}}%
      \csname LTb\endcsname
      \put(1940,559){\rotatebox{45.00}{\makebox(0,0)[r]{\strut{}10.0}}}%
      \csname LTb\endcsname
      \put(2116,677){\makebox(0,0)[l]{\strut{}19.8}}%
      \csname LTb\endcsname
      \put(2116,855){\makebox(0,0)[l]{\strut{}20}}%
      \csname LTb\endcsname
      \put(2116,1033){\makebox(0,0)[l]{\strut{}20.2}}%
      \csname LTb\endcsname
      \put(2116,1211){\makebox(0,0)[l]{\strut{}20.4}}%
      \csname LTb\endcsname
      \put(2116,1389){\makebox(0,0)[l]{\strut{}20.6}}%
      \csname LTb\endcsname
      \put(2116,1567){\makebox(0,0)[l]{\strut{}20.8}}%
    }%
    \gplgaddtomacro\gplfronttext{%
      \csname LTb\endcsname
      \put(127,1077){\rotatebox{-270.00}{\makebox(0,0){\strut{}ROUGE-2}}}%
      \csname LTb\endcsname
      \put(2407,1077){\rotatebox{-270.00}{\makebox(0,0){\strut{}BLEU}}}%
      \csname LTb\endcsname
      \put(1259,351){\makebox(0,0){\strut{}$\beta$}}%
      \csname LTb\endcsname
      \put(1259,1927){\makebox(0,0){\strut{}\footnotesize SPC Dataset}}%
    }%
    \gplbacktext
    \put(0,0){\includegraphics[width={127.00bp},height={99.00bp}]{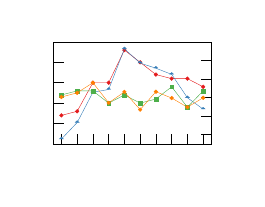}}%
    \gplfronttext
  \end{picture}%
\endgroup

%% file: figures/bw_trend.tex
\begin{figure}[ht]
    \vspace{-5pt}
    \centering

    \begin{subfigure}[ht]{0.49\linewidth}
        {\tiny






        \input{gpl_bw_trend_ed.tex}
        }
    \end{subfigure}%
    \hfill
    \begin{subfigure}[ht]{0.49\linewidth}
        {\tiny






        \input{gpl_bw_trend_spc.tex}
        }
    \end{subfigure}
    
    \vspace{-105pt}
    \begin{tikzpicture}
        \begin{axis}[
            hide axis,
            xmin=0, xmax=1, ymin=0, ymax=1,
            legend columns=4,
            legend style={
                draw=none,
                /tikz/every even column/.append style={column sep=0.5em},
                font=\tiny,
                inner sep=1pt,
            },
        ]
        \addlegendimage{color=c1, mark=*, mark size=1pt, line width=0.8pt}
        \addlegendentry{ROUGE-2 \ourmethodpost}
        \addlegendimage{color=c2, mark=triangle*, mark size=1pt, line width=0.8pt}
        \addlegendentry{BLEU \ourmethodpost}
        \end{axis}
    \end{tikzpicture}
    \vspace{60pt}
    \caption{\ourmethodpost Performance w.r.t. $\alpha$}
    \vspace{-5pt}
    \label{fig:bw_trend}
\end{figure}
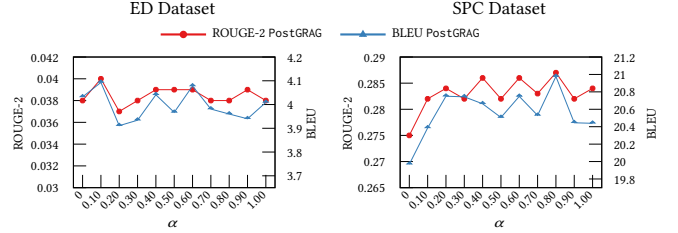

%% file: gpl_bw_trend_ed.tex
\begingroup
  \makeatletter
  \providecommand\color[2][]{%
    \GenericError{(gnuplot) \space\space\space\@spaces}{%
      Package color not loaded in conjunction with
      terminal option `colourtext'%
    }{See the gnuplot documentation for explanation.%
    }{Either use 'blacktext' in gnuplot or load the package
      color.sty in LaTeX.}%
    \renewcommand\color[2][]{}%
  }%
  \providecommand\includegraphics[2][]{%
    \GenericError{(gnuplot) \space\space\space\@spaces}{%
      Package graphicx or graphics not loaded%
    }{See the gnuplot documentation for explanation.%
    }{The gnuplot epslatex terminal needs graphicx.sty or graphics.sty.}%
    \renewcommand\includegraphics[2][]{}%
  }%
  \providecommand\rotatebox[2]{#2}%
  \@ifundefined{ifGPcolor}{%
    \newif\ifGPcolor
    \GPcolortrue
  }{}%
  \@ifundefined{ifGPblacktext}{%
    \newif\ifGPblacktext
    \GPblacktexttrue
  }{}%
  \let\gplgaddtomacro\g@addto@macro
  \gdef\gplbacktext{}%
  \gdef\gplfronttext{}%
  \makeatother
  \ifGPblacktext
    \def\colorrgb#1{}%
    \def\colorgray#1{}%
  \else
    \ifGPcolor
      \def\colorrgb#1{\color[rgb]{#1}}%
      \def\colorgray#1{\color[gray]{#1}}%
      \expandafter\def\csname LTw\endcsname{\color{white}}%
      \expandafter\def\csname LTb\endcsname{\color{black}}%
      \expandafter\def\csname LTa\endcsname{\color{black}}%
      \expandafter\def\csname LT0\endcsname{\color[rgb]{1,0,0}}%
      \expandafter\def\csname LT1\endcsname{\color[rgb]{0,1,0}}%
      \expandafter\def\csname LT2\endcsname{\color[rgb]{0,0,1}}%
      \expandafter\def\csname LT3\endcsname{\color[rgb]{1,0,1}}%
      \expandafter\def\csname LT4\endcsname{\color[rgb]{0,1,1}}%
      \expandafter\def\csname LT5\endcsname{\color[rgb]{1,1,0}}%
      \expandafter\def\csname LT6\endcsname{\color[rgb]{0,0,0}}%
      \expandafter\def\csname LT7\endcsname{\color[rgb]{1,0.3,0}}%
      \expandafter\def\csname LT8\endcsname{\color[rgb]{0.5,0.5,0.5}}%
    \else
      \def\colorrgb#1{\color{black}}%
      \def\colorgray#1{\color[gray]{#1}}%
      \expandafter\def\csname LTw\endcsname{\color{white}}%
      \expandafter\def\csname LTb\endcsname{\color{black}}%
      \expandafter\def\csname LTa\endcsname{\color{black}}%
      \expandafter\def\csname LT0\endcsname{\color{black}}%
      \expandafter\def\csname LT1\endcsname{\color{black}}%
      \expandafter\def\csname LT2\endcsname{\color{black}}%
      \expandafter\def\csname LT3\endcsname{\color{black}}%
      \expandafter\def\csname LT4\endcsname{\color{black}}%
      \expandafter\def\csname LT5\endcsname{\color{black}}%
      \expandafter\def\csname LT6\endcsname{\color{black}}%
      \expandafter\def\csname LT7\endcsname{\color{black}}%
      \expandafter\def\csname LT8\endcsname{\color{black}}%
    \fi
  \fi
    \setlength{\unitlength}{0.0500bp}%
    \ifx\gptboxheight\undefined%
      \newlength{\gptboxheight}%
      \newlength{\gptboxwidth}%
      \newsavebox{\gptboxtext}%
    \fi%
    \setlength{\fboxrule}{0.5pt}%
    \setlength{\fboxsep}{1pt}%
    \definecolor{tbcol}{rgb}{1,1,1}%
\begin{picture}(2540.00,1980.00)%
    \gplgaddtomacro\gplbacktext{%
      \csname LTb\endcsname
      \put(403,588){\makebox(0,0)[r]{\strut{}0.03}}%
      \csname LTb\endcsname
      \put(403,751){\makebox(0,0)[r]{\strut{}0.032}}%
      \csname LTb\endcsname
      \put(403,914){\makebox(0,0)[r]{\strut{}0.034}}%
      \csname LTb\endcsname
      \put(403,1078){\makebox(0,0)[r]{\strut{}0.036}}%
      \csname LTb\endcsname
      \put(403,1241){\makebox(0,0)[r]{\strut{}0.038}}%
      \csname LTb\endcsname
      \put(403,1404){\makebox(0,0)[r]{\strut{}0.04}}%
      \csname LTb\endcsname
      \put(403,1567){\makebox(0,0)[r]{\strut{}0.042}}%
      \csname LTb\endcsname
      \put(572,559){\rotatebox{45.00}{\makebox(0,0)[r]{\strut{}0}}}%
      \csname LTb\endcsname
      \put(710,559){\rotatebox{45.00}{\makebox(0,0)[r]{\strut{}0.10}}}%
      \csname LTb\endcsname
      \put(847,559){\rotatebox{45.00}{\makebox(0,0)[r]{\strut{}0.20}}}%
      \csname LTb\endcsname
      \put(985,559){\rotatebox{45.00}{\makebox(0,0)[r]{\strut{}0.30}}}%
      \csname LTb\endcsname
      \put(1122,559){\rotatebox{45.00}{\makebox(0,0)[r]{\strut{}0.40}}}%
      \csname LTb\endcsname
      \put(1260,559){\rotatebox{45.00}{\makebox(0,0)[r]{\strut{}0.50}}}%
      \csname LTb\endcsname
      \put(1397,559){\rotatebox{45.00}{\makebox(0,0)[r]{\strut{}0.60}}}%
      \csname LTb\endcsname
      \put(1534,559){\rotatebox{45.00}{\makebox(0,0)[r]{\strut{}0.70}}}%
      \csname LTb\endcsname
      \put(1672,559){\rotatebox{45.00}{\makebox(0,0)[r]{\strut{}0.80}}}%
      \csname LTb\endcsname
      \put(1809,559){\rotatebox{45.00}{\makebox(0,0)[r]{\strut{}0.90}}}%
      \csname LTb\endcsname
      \put(1947,559){\rotatebox{45.00}{\makebox(0,0)[r]{\strut{}1.00}}}%
      \csname LTb\endcsname
      \put(2116,677){\makebox(0,0)[l]{\strut{}3.7}}%
      \csname LTb\endcsname
      \put(2116,855){\makebox(0,0)[l]{\strut{}3.8}}%
      \csname LTb\endcsname
      \put(2116,1033){\makebox(0,0)[l]{\strut{}3.9}}%
      \csname LTb\endcsname
      \put(2116,1211){\makebox(0,0)[l]{\strut{}4}}%
      \csname LTb\endcsname
      \put(2116,1389){\makebox(0,0)[l]{\strut{}4.1}}%
      \csname LTb\endcsname
      \put(2116,1567){\makebox(0,0)[l]{\strut{}4.2}}%
    }%
    \gplgaddtomacro\gplfronttext{%
      \csname LTb\endcsname
      \put(107,1077){\rotatebox{-270.00}{\makebox(0,0){\strut{}ROUGE-2}}}%
      \csname LTb\endcsname
      \put(2306,1077){\rotatebox{-270.00}{\makebox(0,0){\strut{}BLEU}}}%
      \csname LTb\endcsname
      \put(1259,327){\makebox(0,0){\strut{}$\alpha$}}%
      \csname LTb\endcsname
      \put(1259,1927){\makebox(0,0){\strut{}\footnotesize ED Dataset}}%
    }%
    \gplbacktext
    \put(0,0){\includegraphics[width={127.00bp},height={99.00bp}]{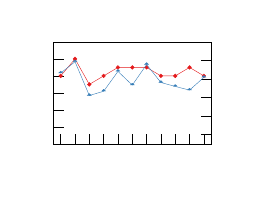}}%
    \gplfronttext
  \end{picture}%
\endgroup

%% file: gpl_bw_trend_spc.tex
\begingroup
  \makeatletter
  \providecommand\color[2][]{%
    \GenericError{(gnuplot) \space\space\space\@spaces}{%
      Package color not loaded in conjunction with
      terminal option `colourtext'%
    }{See the gnuplot documentation for explanation.%
    }{Either use 'blacktext' in gnuplot or load the package
      color.sty in LaTeX.}%
    \renewcommand\color[2][]{}%
  }%
  \providecommand\includegraphics[2][]{%
    \GenericError{(gnuplot) \space\space\space\@spaces}{%
      Package graphicx or graphics not loaded%
    }{See the gnuplot documentation for explanation.%
    }{The gnuplot epslatex terminal needs graphicx.sty or graphics.sty.}%
    \renewcommand\includegraphics[2][]{}%
  }%
  \providecommand\rotatebox[2]{#2}%
  \@ifundefined{ifGPcolor}{%
    \newif\ifGPcolor
    \GPcolortrue
  }{}%
  \@ifundefined{ifGPblacktext}{%
    \newif\ifGPblacktext
    \GPblacktexttrue
  }{}%
  \let\gplgaddtomacro\g@addto@macro
  \gdef\gplbacktext{}%
  \gdef\gplfronttext{}%
  \makeatother
  \ifGPblacktext
    \def\colorrgb#1{}%
    \def\colorgray#1{}%
  \else
    \ifGPcolor
      \def\colorrgb#1{\color[rgb]{#1}}%
      \def\colorgray#1{\color[gray]{#1}}%
      \expandafter\def\csname LTw\endcsname{\color{white}}%
      \expandafter\def\csname LTb\endcsname{\color{black}}%
      \expandafter\def\csname LTa\endcsname{\color{black}}%
      \expandafter\def\csname LT0\endcsname{\color[rgb]{1,0,0}}%
      \expandafter\def\csname LT1\endcsname{\color[rgb]{0,1,0}}%
      \expandafter\def\csname LT2\endcsname{\color[rgb]{0,0,1}}%
      \expandafter\def\csname LT3\endcsname{\color[rgb]{1,0,1}}%
      \expandafter\def\csname LT4\endcsname{\color[rgb]{0,1,1}}%
      \expandafter\def\csname LT5\endcsname{\color[rgb]{1,1,0}}%
      \expandafter\def\csname LT6\endcsname{\color[rgb]{0,0,0}}%
      \expandafter\def\csname LT7\endcsname{\color[rgb]{1,0.3,0}}%
      \expandafter\def\csname LT8\endcsname{\color[rgb]{0.5,0.5,0.5}}%
    \else
      \def\colorrgb#1{\color{black}}%
      \def\colorgray#1{\color[gray]{#1}}%
      \expandafter\def\csname LTw\endcsname{\color{white}}%
      \expandafter\def\csname LTb\endcsname{\color{black}}%
      \expandafter\def\csname LTa\endcsname{\color{black}}%
      \expandafter\def\csname LT0\endcsname{\color{black}}%
      \expandafter\def\csname LT1\endcsname{\color{black}}%
      \expandafter\def\csname LT2\endcsname{\color{black}}%
      \expandafter\def\csname LT3\endcsname{\color{black}}%
      \expandafter\def\csname LT4\endcsname{\color{black}}%
      \expandafter\def\csname LT5\endcsname{\color{black}}%
      \expandafter\def\csname LT6\endcsname{\color{black}}%
      \expandafter\def\csname LT7\endcsname{\color{black}}%
      \expandafter\def\csname LT8\endcsname{\color{black}}%
    \fi
  \fi
    \setlength{\unitlength}{0.0500bp}%
    \ifx\gptboxheight\undefined%
      \newlength{\gptboxheight}%
      \newlength{\gptboxwidth}%
      \newsavebox{\gptboxtext}%
    \fi%
    \setlength{\fboxrule}{0.5pt}%
    \setlength{\fboxsep}{1pt}%
    \definecolor{tbcol}{rgb}{1,1,1}%
\begin{picture}(2540.00,1980.00)%
    \gplgaddtomacro\gplbacktext{%
      \csname LTb\endcsname
      \put(403,588){\makebox(0,0)[r]{\strut{}0.265}}%
      \csname LTb\endcsname
      \put(403,784){\makebox(0,0)[r]{\strut{}0.27}}%
      \csname LTb\endcsname
      \put(403,980){\makebox(0,0)[r]{\strut{}0.275}}%
      \csname LTb\endcsname
      \put(403,1175){\makebox(0,0)[r]{\strut{}0.28}}%
      \csname LTb\endcsname
      \put(403,1371){\makebox(0,0)[r]{\strut{}0.285}}%
      \csname LTb\endcsname
      \put(403,1567){\makebox(0,0)[r]{\strut{}0.29}}%
      \csname LTb\endcsname
      \put(572,559){\rotatebox{45.00}{\makebox(0,0)[r]{\strut{}0}}}%
      \csname LTb\endcsname
      \put(710,559){\rotatebox{45.00}{\makebox(0,0)[r]{\strut{}0.10}}}%
      \csname LTb\endcsname
      \put(847,559){\rotatebox{45.00}{\makebox(0,0)[r]{\strut{}0.20}}}%
      \csname LTb\endcsname
      \put(985,559){\rotatebox{45.00}{\makebox(0,0)[r]{\strut{}0.30}}}%
      \csname LTb\endcsname
      \put(1122,559){\rotatebox{45.00}{\makebox(0,0)[r]{\strut{}0.40}}}%
      \csname LTb\endcsname
      \put(1260,559){\rotatebox{45.00}{\makebox(0,0)[r]{\strut{}0.50}}}%
      \csname LTb\endcsname
      \put(1397,559){\rotatebox{45.00}{\makebox(0,0)[r]{\strut{}0.60}}}%
      \csname LTb\endcsname
      \put(1534,559){\rotatebox{45.00}{\makebox(0,0)[r]{\strut{}0.70}}}%
      \csname LTb\endcsname
      \put(1672,559){\rotatebox{45.00}{\makebox(0,0)[r]{\strut{}0.80}}}%
      \csname LTb\endcsname
      \put(1809,559){\rotatebox{45.00}{\makebox(0,0)[r]{\strut{}0.90}}}%
      \csname LTb\endcsname
      \put(1947,559){\rotatebox{45.00}{\makebox(0,0)[r]{\strut{}1.00}}}%
      \csname LTb\endcsname
      \put(2116,653){\makebox(0,0)[l]{\strut{}19.8}}%
      \csname LTb\endcsname
      \put(2116,784){\makebox(0,0)[l]{\strut{}20}}%
      \csname LTb\endcsname
      \put(2116,914){\makebox(0,0)[l]{\strut{}20.2}}%
      \csname LTb\endcsname
      \put(2116,1045){\makebox(0,0)[l]{\strut{}20.4}}%
      \csname LTb\endcsname
      \put(2116,1175){\makebox(0,0)[l]{\strut{}20.6}}%
      \csname LTb\endcsname
      \put(2116,1306){\makebox(0,0)[l]{\strut{}20.8}}%
      \csname LTb\endcsname
      \put(2116,1437){\makebox(0,0)[l]{\strut{}21}}%
      \csname LTb\endcsname
      \put(2116,1567){\makebox(0,0)[l]{\strut{}21.2}}%
    }%
    \gplgaddtomacro\gplfronttext{%
      \csname LTb\endcsname
      \put(127,1077){\rotatebox{-270.00}{\makebox(0,0){\strut{}ROUGE-2}}}%
      \csname LTb\endcsname
      \put(2407,1077){\rotatebox{-270.00}{\makebox(0,0){\strut{}BLEU}}}%
      \csname LTb\endcsname
      \put(1259,327){\makebox(0,0){\strut{}$\alpha$}}%
      \csname LTb\endcsname
      \put(1259,1927){\makebox(0,0){\strut{}\footnotesize SPC Dataset}}%
    }%
    \gplbacktext
    \put(0,0){\includegraphics[width={127.00bp},height={99.00bp}]{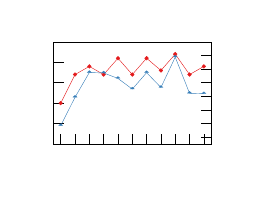}}%
    \gplfronttext
  \end{picture}%
\endgroup

%% file: sections/conclusion.tex
\section{Limitations}
\label{sec:limitations}
While \ourmethod demonstrate strong performance, we note several limitations it faces. 
First, while the goal is to fine-tune and adopt small task LLMs, the intermediate step of generating generic responses still requires a forward pass through the large generic LLM, which makes overall inference process slower than one-step generation using only one model. This issue can be mitigated by leveraging RAG systems~\cite{lewis2020rag} to retrieve the similar utterance and generic response pairs, or by caching frequently occurring utterance-response pairs to improve inference efficiency in real-world deployments.
Second, \ourmethod does not explicitly detect noise in the generic responses. While generic LLMs can introduce valuable semantic grounding signals and potentially useful external knowledge (e.g., Example 3 in Table~\ref{table:generic_example}), they can also introduce detrimental noise into the fine-tuned \ourmethod models via hallucinations or misinformation.
Third, our evaluation focuses on static, text-based  persona datasets. Future work should explore how \ourmethod adapts to dynamic and multimodal user profiles constructed from multiple auxiliary sources over time.

\section{Conclusions}
\label{sec:conclusions}
In this work, we presented a novel framework, Generic Response-Augmented Generation (\ourmethod),  designed to unlock the potential of smaller, task-specific LLMs for personalized conversational  systems in resource-constrained settings. Our approach leverages  generic responses generated by large, general-purpose LLMs as a structural and semantic scaffold. By introducing two architectural variations, \ourmethodpre and \ourmethodpost, we demonstrated that the fundamental bottleneck in personalized conversational systems is often not a lack of persona adherence, but a failure in contextual grounding.

Our experiments and analyses demonstrate  that \ourmethod consistently outperforms  state-of-the-art methods across three benchmark datasets, validating  the effectiveness of leveraging a structural scaffold to keep responses contextually precise while adhering to user profiles. By decoupling high-level discourse structure from personalization, \ourmethod  effectively bridges  the performance gap between smaller edge-device models and massive cloud-based LLMs. Furthermore, it offers  a scalable, privacy-preserving blueprint for deploying sophisticated conversational agents in sensitive or resource-limited environments.

Moving forward, we aim to address the limitations  faced by \ourmethod,  extending  the framework to handle dynamic, interactive, and multimodal user profiles, while exploring  how various task-specific auxiliary signals can guide the generative process. We believe this ``scaffolding'' approach offers a generalized strategy for improving the reliability and performance of smaller language models across broader generative domains.

%% file: sections/appendix.tex
\section*{Appendices}





\section{Prompts}
\label{appendix:prompts}
\begin{figure}[H]
    \centering
    \includegraphics[width=\columnwidth]{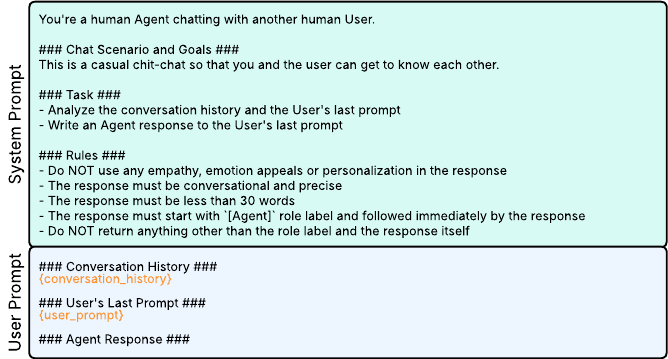}
    \vspace{-15pt}
    \caption{Generic Response Augmentation Prompts - SPC}
    \label{fig:generic_prompts}
    \vspace{-5pt}
\end{figure}

\begin{figure}[H]
    \centering
    \includegraphics[width=\columnwidth]{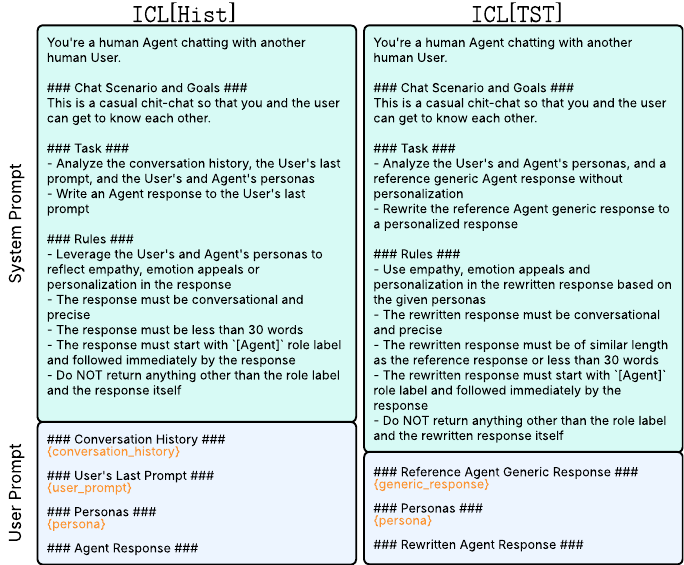}
    \vspace{-15pt}
    \caption{Few-shot ICL Prompts - SPC}
    \label{fig:icl_prompts}
    \vspace{-5pt}
\end{figure}

Figure~\ref{fig:generic_prompts} and~\ref{fig:icl_prompts} show the prompts used in the generic response augmentation and ICL experiments, respectively, for the SPC dataset. The prompts for other dataset are constructed similarly but with different ``Chat Scenario and Goals'' descriptions. In the user prompts, texts in ``\textcolor{orange}{\texttt{\{\}}}'' (e.g., ``\textcolor{orange}{\texttt{\{user\_prompt\}}}'') are replaced with actual data of the conversation turns.

\vspace{-10pt}

%% file: main.bib
@inproceedings{jang2022call,
  title={Call for Customized Conversation: Customized Conversation Grounding Persona and Knowledge},
  author={Jang, Yoonna and Lim, Jungwoo and Hur, Yuna and Oh, Dongsuk and Son, Suhyune and Lee, Yeonsoo and Shin, Donghoon and Kim, Seungryong and Lim, Heuiseok},
  booktitle={Proceedings of the AAAI Conference on Artificial Intelligence},
  volume={36},
  number={10},
  pages={10803--10812},
  year={2022}
}

@inproceedings{lewis2019bart,
  title={BART: Denoising sequence-to-sequence pre-training for natural language generation, translation, and comprehension},
  author={Lewis, Mike and Liu, Yinhan and Goyal, Naman and Ghazvininejad, Marjan and Mohamed, Abdelrahman and Levy, Omer and Stoyanov, Veselin and Zettlemoyer, Luke},
  booktitle={Proceedings of the 58th annual meeting of the association for computational linguistics},
  pages={7871--7880},
  year={2020}
}

@article{radford2019gpt2,
  title={Language models are unsupervised multitask learners},
  author={Radford, Alec and Wu, Jeffrey and Child, Rewon and Luan, David and Amodei, Dario and Sutskever, Ilya and others},
  journal={OpenAI blog},
  volume={1},
  number={8},
  pages={9},
  year={2019}
}

@inproceedings{zhang2018personalizing,
  title={Personalizing dialogue agents: I have a dog, do you have pets too?},
  author={Zhang, Saizheng and Dinan, Emily and Urbanek, Jack and Szlam, Arthur and Kiela, Douwe and Weston, Jason},
  booktitle={Proceedings of the 56th Annual Meeting of the Association for Computational Linguistics (Volume 1: Long Papers)},
  pages={2204--2213},
  year={2018}
}

@inproceedings{humeau2019poly,
  title={Poly-encoders: Architectures and Pre-training Strategies for Fast and Accurate Multi-sentence Scoring},
  author={Humeau, Samuel and Shuster, Kurt and Lachaux, Marie-Anne and Weston, Jason},
  booktitle={International Conference on Learning Representations},
  year={2019}
}

@inproceedings{hoogeboom2021autoregressive,
  title={Autoregressive Diffusion Models},
  author={Hoogeboom, Emiel and Gritsenko, Alexey A and Bastings, Jasmijn and Poole, Ben and van den Berg, Rianne and Salimans, Tim},
  booktitle={International Conference on Learning Representations},
  year={2021}
}

@inproceedings{dinan2018wizard,
  author={Emily Dinan and Stephen Roller and Kurt Shuster and Angela Fan and Michael Auli and Jason Weston},
  title={{W}izard of {W}ikipedia: Knowledge-powered Conversational Agents},
  booktitle = {Proceedings of the International Conference on Learning Representations (ICLR)},
  year={2019},
}

@inproceedings{liu2022persona,
  title={Persona-Based Conversational AI: State of the Art and Challenges},
  author={Liu, Junfeng and Symons, Christopher and Vatsavai, Ranga Raju},
  booktitle={2022 IEEE International Conference on Data Mining Workshops (ICDMW)},
  pages={993--1001},
  year={2022},
  organization={IEEE}
}

@inproceedings{liu2023pcpe,
  title={Persona-coded poly-encoder: persona-guided multi-stream conversational sentence scoring},
  author={Liu, Junfeng and Symons, Christopher and Vatsavai, Ranga Raju},
  booktitle={2023 IEEE 35th International Conference on Tools with Artificial Intelligence (ICTAI)},
  pages={250--257},
  year={2023},
  organization={IEEE}
}

@inproceedings{mostafazadeh2017image,
  title={Image-grounded conversations: Multimodal context for natural question and response generation},
  author={Mostafazadeh, Nasrin and Brockett, Chris and Dolan, William B and Galley, Michel and Gao, Jianfeng and Spithourakis, Georgios and Vanderwende, Lucy},
  booktitle={Proceedings of the Eighth International Joint Conference on Natural Language Processing (Volume 1: Long Papers)},
  pages={462--472},
  year={2017}
}

@inproceedings{liu2025comac,
  title={CoMAC: Conversational Agent for Multi-source Auxiliary Context with Sparse and Symmetric Latent Interactions},
  author={Liu, Junfeng and Symons, Christopher T and Vatsavai, Ranga Raju},
  booktitle={Pacific-Asia Conference on Knowledge Discovery and Data Mining},
  pages={196--208},
  year={2025},
  organization={Springer}
}

@incollection{dinan2020second,
  title={The second conversational intelligence challenge (convai2)},
  author={Dinan, Emily and Logacheva, Varvara and Malykh, Valentin and Miller, Alexander and Shuster, Kurt and Urbanek, Jack and Kiela, Douwe and Szlam, Arthur and Serban, Iulian and Lowe, Ryan and others},
  booktitle={The NeurIPS'18 Competition},
  pages={187--208},
  year={2020},
  publisher={Springer}
}

@inproceedings{rashkin2019towards,
  title={Towards empathetic open-domain conversation models: A new benchmark and dataset},
  author={Rashkin, Hannah and Smith, Eric Michael and Li, Margaret and Boureau, Y-Lan},
  booktitle={Proceedings of the 57th Annual Meeting of the Association for Computational Linguistics},
  pages={5370--5381},
  year={2019}
}

@inproceedings{wang2019persuasion,
  title={Persuasion for Good: Towards a Personalized Persuasive Dialogue System for Social Good},
  author={Wang, Xuewei and Shi, Weiyan and Kim, Richard and Oh, Yoojung and Yang, Sijia and Zhang, Jingwen and Yu, Zhou},
  booktitle={Proceedings of the 57th Annual Meeting of the Association for Computational Linguistics},
  pages={5635--5649},
  year={2019}
}

@inproceedings{jandaghi2024faithful,
  title={Faithful persona-based conversational dataset generation with large language models},
  author={Jandaghi, Pegah and Sheng, XiangHai and Bai, Xinyi and Pujara, Jay and Sidahmed, Hakim},
  booktitle={Proceedings of the 6th Workshop on NLP for Conversational AI (NLP4ConvAI 2024)},
  pages={114--139},
  year={2024}
}

@inproceedings{huang2025llmjepa,
  title={LLM-JEPA: Large Language Models Meet Joint Embedding Predictive Architectures},
  author={Huang, Hai and LeCun, Yann and Balestriero, Randall},
  booktitle={NeurIPS 2025 Fourth Workshop on Deep Learning for Code},
  year={2025}
}

@inproceedings{assran2023self,
  title={Self-supervised learning from images with a joint-embedding predictive architecture},
  author={Assran, Mahmoud and Duval, Quentin and Misra, Ishan and Bojanowski, Piotr and Vincent, Pascal and Rabbat, Michael and LeCun, Yann and Ballas, Nicolas},
  booktitle={Proceedings of the IEEE/CVF conference on computer vision and pattern recognition},
  pages={15619--15629},
  year={2023}
}

@article{bucher2023patient,
  title={The patient experience of the future is personalized: using technology to scale an N of 1 approach},
  author={Bucher, Amy},
  journal={Journal of Patient Experience},
  volume={10},
  pages={23743735231167975},
  year={2023},
  publisher={SAGE Publications Sage CA: Los Angeles, CA}
}

@inproceedings{liu2020you,
  title={You impress me: Dialogue generation via mutual persona perception},
  author={Liu, Qian and Chen, Yihong and Chen, Bei and Lou, Jian-Guang and Chen, Zixuan and Zhou, Bin and Zhang, Dongmei},
  booktitle={Proceedings of the 58th annual meeting of the association for computational linguistics},
  pages={1417--1427},
  year={2020}
}

@inproceedings{zhang2020dialogpt,
  title={Dialogpt: Large-scale generative pre-training for conversational response generation},
  author={Zhang, Yizhe and Sun, Siqi and Galley, Michel and Chen, Yen-Chun and Brockett, Chris and Gao, Xiang and Gao, Jianfeng and Liu, Jingjing and Dolan, William B},
  booktitle={Proceedings of the 58th annual meeting of the association for computational linguistics: system demonstrations},
  pages={270--278},
  year={2020}
}

@article{rothe2020leveraging,
  title={Leveraging pre-trained checkpoints for sequence generation tasks},
  author={Rothe, Sascha and Narayan, Shashi and Severyn, Aliaksei},
  journal={Transactions of the Association for Computational Linguistics},
  volume={8},
  pages={264--280},
  year={2020},
  publisher={MIT Press One Rogers Street, Cambridge, MA 02142-1209, USA journals-info~…}
}

@article{wang2023large,
  title={Large language models are latent variable models: Explaining and finding good demonstrations for in-context learning},
  author={Wang, Xinyi and Zhu, Wanrong and Saxon, Michael and Steyvers, Mark and Wang, William Yang},
  journal={Advances in Neural Information Processing Systems},
  volume={36},
  pages={15614--15638},
  year={2023}
}

@inproceedings{zhang2024distilling,
  title={Distilling text style transfer with self-explanation from LLMs},
  author={Zhang, Chiyu and Cai, Honglong and Li, Yuezhang and Wu, Yuexin and Hou, Le and Abdul-Mageed, Muhammad},
  booktitle={Proceedings of the 2024 Conference of the North American Chapter of the Association for Computational Linguistics: Human Language Technologies (Volume 4: Student Research Workshop)},
  pages={200--211},
  year={2024}
}

@inproceedings{reif2022recipe,
  title={A recipe for arbitrary text style transfer with large language models},
  author={Reif, Emily and Ippolito, Daphne and Yuan, Ann and Coenen, Andy and Callison-Burch, Chris and Wei, Jason},
  booktitle={Proceedings of the 60th Annual Meeting of the Association for Computational Linguistics (Volume 2: Short Papers)},
  pages={837--848},
  year={2022}
}

@article{yu2023language,
  title={Language as a latent sequence: Deep latent variable models for semi-supervised paraphrase generation},
  author={Yu, Jialin and Cristea, Alexandra I and Harit, Anoushka and Sun, Zhongtian and Aduragba, Olanrewaju Tahir and Shi, Lei and Al Moubayed, Noura},
  journal={AI Open},
  volume={4},
  pages={19--32},
  year={2023},
  publisher={Elsevier}
}

@article{jia2025syntax,
  title={Syntax-controlled paraphrases generation with VAE and multi-task learning},
  author={Jia, Xiyuan and Mao, Zongqing and Zhang, Zhen and Lv, Qiyun and Wang, Xin and Wu, Guohua},
  journal={Computer Speech \& Language},
  volume={89},
  pages={101705},
  year={2025},
  publisher={Elsevier}
}

@article{pan2024unsupervised,
  title={Unsupervised text style transfer via LLMs and attention masking with multi-way interactions},
  author={Pan, Lei and Lan, Yunshi and Li, Yang and Qian, Weining},
  journal={arXiv preprint arXiv:2402.13647},
  year={2024}
}

@article{hu2022text,
  title={Text style transfer: A review and experimental evaluation},
  author={Hu, Zhiqiang and Lee, Roy Ka-Wei and Aggarwal, Charu C and Zhang, Aston},
  journal={ACM SIGKDD Explorations Newsletter},
  volume={24},
  number={1},
  pages={14--45},
  year={2022},
  publisher={ACM New York, NY, USA}
}

@inproceedings{lample2019multiple,
  title={Multiple-attribute text rewriting},
  author={Lample, Guillaume and Subramanian, Sandeep and Smith, Eric and Denoyer, Ludovic and Ranzato, Marc'Aurelio and Boureau, Y-Lan},
  booktitle={International Conference on Learning Representations},
  year={2019}
}

@inproceedings{hu2017toward,
  title={Toward controlled generation of text},
  author={Hu, Zhiting and Yang, Zichao and Liang, Xiaodan and Salakhutdinov, Ruslan and Xing, Eric P},
  booktitle={International conference on machine learning},
  pages={1587--1596},
  year={2017},
  organization={PMLR}
}

@inproceedings{saha2022stylistic,
  title={Stylistic response generation by controlling personality traits and intent},
  author={Saha, Sougata and Das, Souvik and Srihari, Rohini K},
  booktitle={Proceedings of the 4th Workshop on NLP for Conversational AI},
  pages={197--211},
  year={2022}
}

@article{zhang2019neural,
  title={Neural personalized response generation as domain adaptation},
  author={Zhang, Wei-Nan and Zhu, Qingfu and Wang, Yifa and Zhao, Yanyan and Liu, Ting},
  journal={World Wide Web},
  volume={22},
  number={4},
  pages={1427--1446},
  year={2019},
  publisher={Springer}
}

@inproceedings{zhang2020bertscore,
  title={BERTScore: Evaluating Text Generation with BERT},
  author={Zhang, Tianyi and Kishore, Varsha and Wu, Felix and Weinberger, Kilian Q and Artzi, Yoav},
  booktitle={International Conference on Learning Representations},
  year={2020}
}

@article{lewis2020rag,
  title={Retrieval-augmented generation for knowledge-intensive nlp tasks},
  author={Lewis, Patrick and Perez, Ethan and Piktus, Aleksandra and Petroni, Fabio and Karpukhin, Vladimir and Goyal, Naman and K{\"u}ttler, Heinrich and Lewis, Mike and Yih, Wen-tau and Rockt{\"a}schel, Tim and others},
  journal={Advances in neural information processing systems},
  volume={33},
  pages={9459--9474},
  year={2020}
}

@techreport{achiam2023gpt,
  title={Gpt-4 technical report},
  author={Achiam, Josh and others},
  journal={arXiv preprint arXiv:2303.08774},
  year={2023},
  url={https://arxiv.org/abs/2303.08774}
}

@techreport{anthropic2025claude4,
  title        = {System Card: {Claude Opus 4} \& {Claude Sonnet 4}},
  author       = {{Anthropic, AI}},
  year         = {2025},
  month        = {May},
  institution  = {{Anthropic, AI}},
  url          = {https://www-cdn.anthropic.com/4263b940cabb546aa0e3283f35b686f4f3b2ff47.pdf},
  journal={Claude-4 Model Card},
}

@inproceedings{hu2022lora,
  title={LoRA: Low-Rank Adaptation of Large Language Models},
  author={Hu, Edward J and Wallis, Phillip and Allen-Zhu, Zeyuan and Li, Yuanzhi and Wang, Shean and Wang, Lu and Chen, Weizhu and others},
  booktitle={Proceedings of the International Conference on Learning Representations (ICLR)},
  year={2022}
}

@article{touvron2023llama,
  title={LLaMA: Open and efficient foundation language models.},
  author={Touvron, Hugo and Lavril, Thibaut and Izacard, Gautier and Martinet, Xavier and Lachaux, Marie-Anne and Lacroix, Timoth{\'e}e and Rozi{\`e}re, Baptiste and Goyal, Naman and Hambro, Eric and Azhar, Faisal and others},
  journal={arXiv preprint arXiv:2302.13971},
  volume={10},
  year={2023}
}

@inproceedings{roller2021recipes,
  title={Recipes for building an open-domain chatbot},
  author={Roller, Stephen and Dinan, Emily and Goyal, Naman and Ju, Da and Williamson, Mary and Liu, Yinhan and Xu, Jing and Ott, Myle and Smith, Eric Michael and Boureau, Y-Lan and others},
  booktitle={Proceedings of the 16th Conference of the European Chapter of the Association for Computational Linguistics: Main Volume},
  pages={300--325},
  year={2021}
}

@inproceedings{dong2024survey,
  title={A survey on in-context learning},
  author={Dong, Qingxiu and Li, Lei and Dai, Damai and Zheng, Ce and Ma, Jingyuan and Li, Rui and Xia, Heming and Xu, Jingjing and Wu, Zhiyong and Chang, Baobao and others},
  booktitle={Proceedings of the 2024 conference on empirical methods in natural language processing},
  pages={1107--1128},
  year={2024}
}

@article{hinton2015distilling,
  title={Distilling the knowledge in a neural network},
  author={Hinton, Geoffrey and Vinyals, Oriol and Dean, Jeff},
  journal={arXiv preprint arXiv:1503.02531},
  year={2015}
}

@inproceedings{dai2019style,
  title={Style transformer: Unpaired text style transfer without disentangled latent representation},
  author={Dai, Ning and Liang, Jianze and Qiu, Xipeng and Huang, Xuan-Jing},
  booktitle={Proceedings of the 57th annual meeting of the association for computational linguistics},
  pages={5997--6007},
  year={2019}
}

@article{chen2025vl-jepa,
  title={VL-JEPA: Joint embedding predictive architecture for vision-language},
  author={Chen, Delong and Shukor, Mustafa and Moutakanni, Theo and Chung, Willy and Yu, Jade and Kasarla, Tejaswi and Bang, Yejin and Bolourchi, Allen and LeCun, Yann and Fung, Pascale},
  journal={arXiv preprint arXiv:2512.10942},
  year={2025}
}

@article{raffel2020t5,
  title={Exploring the limits of transfer learning with a unified text-to-text transformer},
  author={Raffel, Colin and Shazeer, Noam and Roberts, Adam and Lee, Katherine and Narang, Sharan and Matena, Michael and Zhou, Yanqi and Li, Wei and Liu, Peter J},
  journal={Journal of machine learning research},
  volume={21},
  number={140},
  pages={1--67},
  year={2020}
}

@article{yang2025qwen3,
  title={Qwen3 technical report},
  author={Yang, An and Li, Anfeng and Yang, Baosong and Zhang, Beichen and Hui, Binyuan and Zheng, Bo and Yu, Bowen and Gao, Chang and Huang, Chengen and Lv, Chenxu and others},
  journal={arXiv preprint arXiv:2505.09388},
  year={2025}
}

@article{dettmers2023qlora,
  title={QLoRA: Efficient finetuning of quantized llms},
  author={Dettmers, Tim and Pagnoni, Artidoro and Holtzman, Ari and Zettlemoyer, Luke},
  journal={Advances in neural information processing systems},
  volume={36},
  pages={10088--10115},
  year={2023}
}

@inproceedings{loshchilov2019decoupled,
  title={Decoupled Weight Decay Regularization},
  author={Loshchilov, Ilya and Hutter, Frank},
  booktitle={Proceedings of the International Conference on Learning Representations (ICLR)},
  year={2019}
}
